\documentclass{article}
\usepackage{arxiv}

\usepackage{amsmath}
\usepackage{amsfonts}
\usepackage[capitalize,noabbrev]{cleveref}
\usepackage{algorithmic}
\usepackage{graphicx}
\usepackage{textcomp}
\usepackage{xcolor}

\usepackage{booktabs} 
\usepackage{multirow}
\usepackage{xspace}
\usepackage{color}
\usepackage{balance}
\usepackage{url}
\usepackage{subcaption}
\usepackage{pgf}
\usepackage[ruled,linesnumbered]{algorithm2e}
\usepackage[normalem]{ulem}
\usepackage[font=footnotesize]{subcaption}

\newcommand{\name}{Arachne\xspace}
\newcommand{\corig}{\ensuremath{c_{original}}\xspace}
\newcommand{\cpred}{\ensuremath{c_{predicted}}\xspace}

\newcommand{\ns}{Arac.\xspace}
\newcommand{\model}{\ensuremath{\mathbf{M}}}
\newcommand{\loss}{\ensuremath{\mathbf{\mathcal{L}}}}
\newcommand{\new}[1]{{\color{black}#1}}

\newcommand{\minorrev}[1]{{\color{black}#1}}

\newcommand*{\Numptwo}[1]{%
    \pgfmathprintnumber[
        fixed,
        precision=2,
        fixed zerofill=true,
        ]{#1}}

\newcommand*{\Numpthree}[1]{%
    \pgfmathprintnumber[
        fixed,
        precision=3,
        fixed zerofill=true,
        ]{#1}}

\newcommand*{\Numpfour}[1]{%
    \pgfmathprintnumber[
        fixed,
        precision=4,
        fixed zerofill=true,
        ]{#1}}

\def\BibTeX{{\rm B\kern-.05em{\sc i\kern-.025em b}\kern-.08em
    T\kern-.1667em\lower.7ex\hbox{E}\kern-.125emX}}

\title{\name: Search Based Repair of Deep Neural Networks}
\date{}

\author{
Jeongju Sohn\thanks{First co-authors with equal contribution}\\
\small University of Luxembourg\\
\small Luxembourg city\\
\small Luxembourg\\
\small \texttt{jeongju.sohn@uni.lu}
\And
Sungmin Kang\footnotemark[1]\\
\small KAIST\\
\small Daejeon\\
\small Republic of Korea\\
\small \texttt{sungmin.kang@kaist.ac.kr}
\And
Shin Yoo\\
\small KAIST\\
\small Daejeon\\
\small Republic of Korea\\
\small \texttt{shin.yoo@kaist.ac.kr}
}

\begin{document}
\maketitle

\begin{abstract}

The rapid and widespread adoption of Deep Neural Networks (DNNs) has called for ways to test their behaviour, and many testing approaches have successfully revealed misbehaviour of DNNs. However, it is relatively unclear what one can do to correct such behaviour after revelation, as retraining involves costly data collection and does not guarantee to fix the underlying issue. This paper introduces \name, a novel program repair technique for DNNs, which directly repairs DNNs using their input-output pairs as a specification. \name localises neural weights on which it can generate effective patches and uses Differential Evolution to optimise the localised weights and correct the misbehaviour. An empirical study using different benchmarks shows that \name can fix specific misclassifications of a DNN without reducing general accuracy significantly. On average, patches generated by \name generalise to \new{61.3\%} of unseen misbehaviour, whereas those by a state-of-the-art DNN repair technique generalise only to \new{10.2\% and sometimes to none while taking tens of times more than \name.} We also show that \name can address fairness issues by debiasing a gender classification model. \new{Finally, we successfully apply \name to a text sentiment model to show that it generalises beyond Convolutional Neural Networks.}

\end{abstract}

\section{Introduction}
\label{sec:intro}

Deep Neural Networks (DNNs) are rapidly being adopted in many 
application areas~\cite{LeCun2015ef}, ranging from image recognition~\cite{
Krizhevsky2017aa,Szegedy2015pt}, speech recognition~\cite{6296526}, machine 
translation~\cite{Jean2015aa,Sutskever:2014aa}, to safety-critical domains 
such as autonomous driving~\cite{chen2015deepdriving,Chen2017aa} and medical 
imaging~\cite{Litjens2017rt}. As the application areas expand, there has 
been a growing concern that these DNNs should be \emph{tested}, both in 
isolation and as a part of larger systems, to ensure dependable performance. 
The need for testing resulted in two major classes of techniques. First, to 
evaluate sets of test inputs, many test adequacy criteria have recently been 
proposed~\cite{Pei2017qy,Tian2018aa,Kim2019aa}. Second, ways to 
synthesise new inputs by applying small perturbations to given inputs
(such as emulation of different lighting or weather conditions)  
have been introduced~\cite{Ma2018aa,Zhang2018aa,Kang2020aa}. 
Newly synthesised inputs not only increase 
input diversity, but also can 
reveal unexpected behaviour of DNNs under test. 

Compared to traditional software developed by human engineers, however, 
the stages after the detection of unexpected behaviour remain relatively 
unexplored for DNNs. This is due to the major difference between the way DNNs 
and code are developed: one is trained based on data, 
while the other is written by human engineers based on specifications. 
Consequently, existing efforts to \emph{repair} the unexpected behaviour 
tend to heavily rely on retraining~\cite{Ma2018gf}.
However, the use of retraining as a means to repair a DNN has a couple of 
weaknesses. 
First, since retraining uses the overall accuracy of a DNN to assess the quality of a repair, it may not remove the unexpected behaviour even if retraining increases overall accuracy. 
Apricot, a state-of-the-art DNN repair technique~\cite{Zhang2019aa}, fixes 
DNNs iteratively by incorporating
training into its repairing process. As Apricot also aims to increase general
accuracy, it is not geared towards removing unexpected behaviour and
may have difficulty removing a specific misclassification pair. 
Second, systematic retraining can be 
computationally expensive. In the case of Apricot, there is a training loop
within the repairing process, which causes Apricot to take multiple hours 
repairing DNNs.

This paper introduces \name, a search-based automated program repair (APR) 
technique for DNN classifiers. Compared to existing techniques, \name is 
more focused on fixing specific misclassifications, e.g. erroneously predicting an input of class $a$ as class $b$, rather than overall accuracy.
\name directly searches the space of neural weights,
guided by a novel 
fitness function inspired by Generate and Validate APR techniques~\cite{
Goues:2012zr,Yuan2018zl,Saha2019jh}. \name resembles APR techniques for 
traditional code in many aspects: it attempts to localise components
relevant to the given misclassification, and uses both positive (i.e., inputs 
that are correctly classified) and negative inputs (i.e., inputs that are  
misclassified) to retain correct behaviour and to generate a patch, respectively. 
As \name directly changes the values of neural weights, 
its internal representation of a patch is a vector of real numbers. 
\name thus adopts Differential Evolution (DE)~\cite{Storn1997aa}, a meta-heuristic optimisation algorithm shown to be effective in continuous search spaces~\cite{Storn1997aa,Das2011aa}, as its search algorithm.  
During each fitness evaluation, \name updates the localised neural weights of the 
DNN under repair with values with the DE candidate solutions.
Subsequently, it evaluates the fitness of solutions based on the classification results of inputs.

We empirically evaluate \name on four image classification benchmarks (i.e., Fashion-MNIST~\cite{Xiao2017jy}, CIFAR-10~\cite{Krizhevsky2009kt}, \new{German Traffic Sign Recognition Benchmark (GTSRB)~\cite{Stallkamp:ijcn:2011}} and Labelled Faces in the Wild (LFW)~\cite{LFWTech}) and \new{one text classification benchmark (i.e., Twitter US Airline Sentiment dataset~\cite{us_airline:tweets}).}
We use the first three benchmarks to show the feasibility of fault localisation and patch generation. We also use them to compare \name to Apricot: for CIFAR-10, we reuse the three CNN architectures employed to evaluate Apricot~\cite{Zhang2019aa} for fair comparison \new{and build an additional image classifier model with more training capacity than the one used to evaluate the feasibility individually for Fashion-MNIST and GTSRB.}
The LFW benchmark is an award-winning benchmark of human faces: we use it to construct a case study that shows how \name can fix a fairness issue in a gender classifier by rebalancing the model. \new{For the Twitter benchmark, which contains various tweets and their underlying sentiments, we employ it to evaluate whether \name can address the models other than those for image classification.}

The technical contributions of this paper are as follows:
\begin{itemize} 
    \item The paper introduces \name, a novel search-based repair technique for DNNs. Unlike existing approaches that retrain a DNN model using more
    inputs, \name aims to repair a pre-trained model by directly
    adjusting neural weights. 

    \item We empirically evaluate \name against the state-of-the-art DNN 
    repair technique, Apricot, \new{using widely studied image classification benchmarks.} \name can produce repairs that are more focused on the targeted misclassifications, while only minimally perturbing other behaviour, and operate \new{at speeds tens of times faster than Apricot.}

    \item We present a case study on how \name can be used to repair a DNN 
    model suffering from a fairness issue. \name can successfully repair bias in the given gender classifier without requiring any additional data. 
    \name improved the classification accuracy for female images from 86.8\% to 88.8\%.

    \item \new{We conduct an additional study with a model that analyses the underlying sentiment of text in order to evaluate whether \name can be effective in different domains besides image classification. 
    \name successfully decreases the prevalence of the most frequent errors, showing that \name can work with diverse models.}

\end{itemize}

The remainder of this paper is organised as follows. 
Section~\ref{sec:technique} describes components of our DNN repair technique, \name. 
Section~\ref{sec:protocols} sets out the research questions and 
describes experimental protocols. 
Section~\ref{sec:experimental_setup} outlines the set-up of the empirical evaluation, 
the results of which are discussed in Section~\ref{sec:results}. Section~\ref{sec:threats}
presents threats to validity, Section~\ref{sec:related_work} details the related work, 
and Section~\ref{sec:conclusion} concludes.

\section{\name: Search Based Repair for DNNs}
\label{sec:technique}

This section motivates the development of \name and describes its internal components.

\subsection{Motivation} %
\label{sub:motivation}

The standard approach towards getting rid of misbehaviour in DNNs is 
through further training, which typically involves a large volume of training 
data, curated carefully to avoid the misbehaviour.
Unfortunately, preparing training data is known to be the major bottleneck
to the practical application of machine learning due to its high cost~\cite{Amershi2019aa,Roh2018aa}. Furthermore, even with new curated training data, retraining does not always remove the observed misbehaviour, as retraining aims to improve overall accuracy instead.

\new{
We posit that there are situations that call for \emph{targeted} improvements 
of DNN models with more assurance for the improved outcome. For example, 
certain misclassifications, such as mistaking a stop sign for a 60 kilometers 
per hour speed limit sign, may pose more risk than other mistakes about parking 
allowance. One may also argue that some misbehaviour that stem from inherent
bias in the data should be fixed, even if it degrades overall accuracy, as 
in the case of gender bias in facial classification~\cite{buolamwini2018gender}.
Presented with such misbehaviour that needs to be repaired urgently, 
retraining loses its appeal as a correction mechanism due to the cost of data 
curation. In such a scenario, a more direct approach that does not require 
additional data may be more useful.}

\new{
We emphasise that our aim with \name is not to replace well-designed learning 
processes as a means of improving general learning capability and overall model 
accuracy. Rather, we want to introduce and evaluate an alternative repair 
technique that can introduce a directed and focused improvement over a small 
set of unexpected behaviour, without requiring larger and better-curated 
training datasets. In this sense, we expect \name to complement other 
training-based techniques. Note that \name targets misbehaviour that can be 
repaired by neural weight manipulation alone: if a DNN model underperforms due 
to its inappropriate architecture, we do not expect \name to be effective at 
repairing it.}

\subsection{Overview} %
\label{sub:overview}

\name has two primary operations: localisation and patch generation.
In the localisation phase, 
\name identifies a set of neural weights that are likely to be 
related to the observed misclassification revealed by a negative input set.
The intuition is that changing the values of these neural weights is likely to 
affect the model behaviour in the desired direction.

\new{As an effective repair should fix misbehaviour while minimally disrupting correct behaviour, \name leverages positive inputs (i.e., inputs that are processed correctly) in addition to to negative inputs (i.e., inputs that reveal the misclassification) for both localisation and patch generation. 
The basic intuition behind fault localisation in software engineering is that faulty elements relate more to a program's faulty behaviour (e.g., failing tests) and less to its correct behaviour (e.g., passing tests)~\cite{Wong:2016aa}. Under the same intuition, \name localises neural weights that have more impact on negative inputs and less impact on positive inputs.
In the patch generation phase,} similarly to Generate and Validate (G\&V) Automated Program Repair techniques such as GenProg~\cite{Weimer:2009fk} and Arja~\cite{Yuan2018zl}, the positive inputs are used by the fitness function of \name to retain the initially correct behaviour of the DNN under repair. Employing fitness values based on the inference results of both positive and negative inputs, \name uses Differential Evolution~\cite{Storn1997aa} to search for a set of neural weights that would correct the behaviour of the DNN under repair. 
Details of the localisation and patch generation phases are discussed in the following subsections.

 \begin{figure}[ht!]
 \centering
 \caption{\name fault localisation diagram. 
 Black arrows represent forward pass, while red and orange arrows are backpropagation.
 \new{Black and red arrows combined compute the forward impact, whereas the orange arrow denotes the gradient loss.
 The gradient loss and the forward impact combined evaluate the likelihood of $w_{i,j}$ being localised by \name. 
 }
 \label{fig:target_layer}}
 \includegraphics[width=0.55\textwidth]{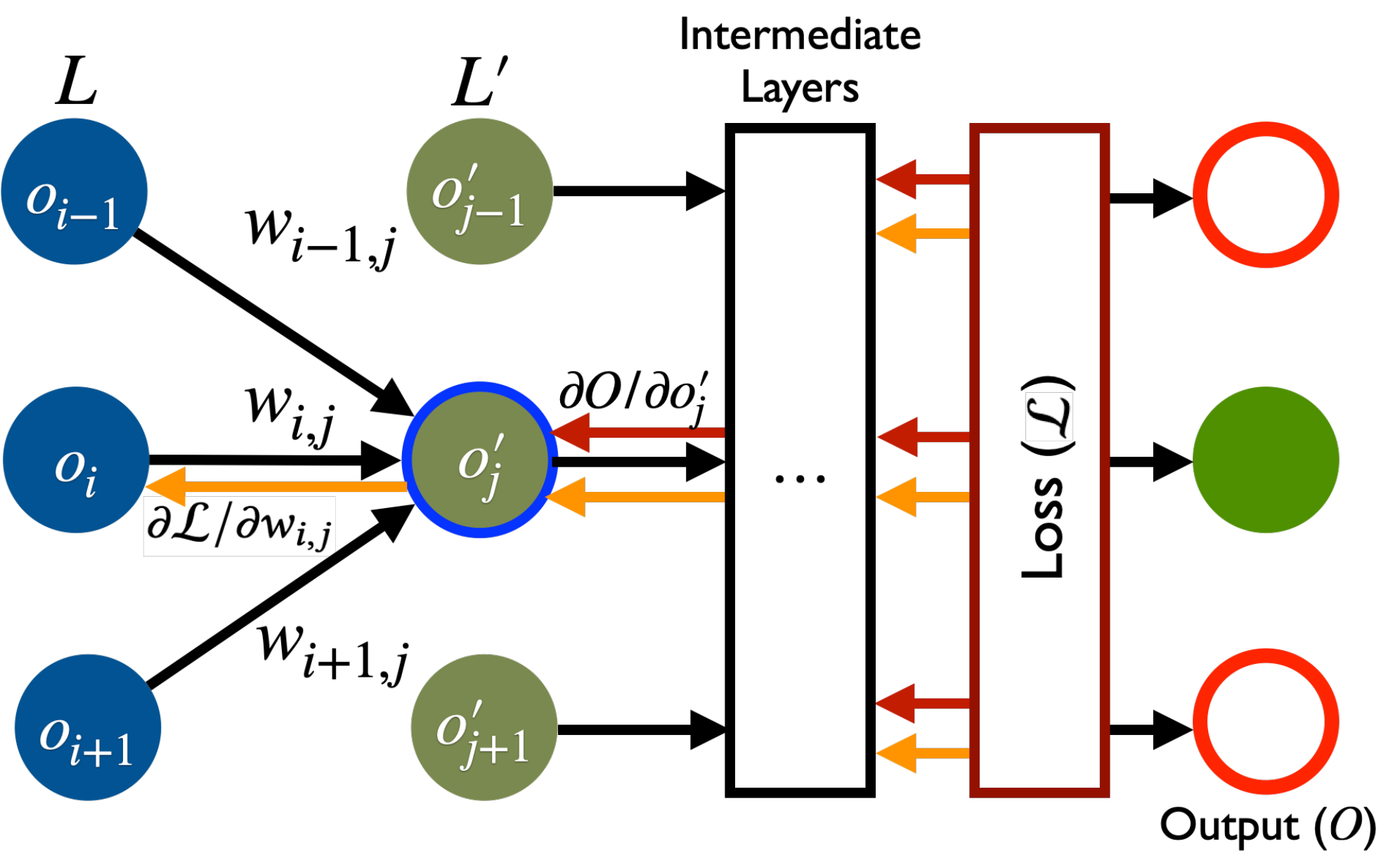}
 \end{figure}

\subsection{Localisation Phase} %
\label{sub:fixing_point_localisation}

\new{Attempting to adjust all neural weights of a DNN would be costly, as even relatively simple DNNs consist of thousands of weight parameters. 
Hence, instead of targeting all weights of the DNN, \name incorporates a novel localisation method, Bidirectional Localisation (BL), that aims to identify neural weights on which \name can more effectively repair. 
As depicted in \cref{fig:target_layer}, BL considers the gradient loss and the forward impact of each neural weight; the former quantifies the responsibility that a neural weight has for the misclassification, and the latter measures the influence of the weight on the final classification outcome. 
Since a weight that has a high impact is not necessarily responsible for the misclassification, BL treats the forward impact and the gradient loss as two competing objectives to optimise.

Algorithm~\ref{alg:loc} presents the pseudo-code of the bidirectional localisation method. The algorithm calls \textbf{ComputeGradientLoss} and \textbf{ComputeForwardImpact} to compute the gradient loss (Line 4-5) and the forward impact (Line 7-8) for each neural weight. Computing the gradient loss of a neural weight is straightforward. 
For instance, for a neural weight $w_{i,j}$ that connects the $i$th neuron in layer $L$, $o_i$, and the $j$th neuron in the following layer $L'$, $o'_j$, in \cref{fig:target_layer}, its gradient loss backpropagated from the final loss, $\loss$, is simply calculated as $\frac{\partial \loss}{\partial w_{i,j}}$. 

Computing the forward impact of a neural weight on the final output (\textbf{ComputeForwardImpact}) is more complicated than computing the gradient loss (\textbf{ComputeGradientLoss}). 
For a neural weight to have a significant impact on the final classification outcome, the weight has to be highly influential on the activation value of the output neuron to which it is connected (1), and this output neuron should be able to influence the final classification result (2). 
BL estimates the influence of a neural weight on the activation of the output neuron by multiplying the given neural weight and the average activation value of the neuron connected to the weight in the previous layer; it further normalises this estimated influence by dividing it by the aggregated influence of the neural weights connected to the same output neuron.
Hence, in \cref{fig:target_layer}, the influence of $w_{i,j}$ on the activation of $o'_{j}$ (1) is computed as $o_{i} w_{i,j}$, which is then normalised as $\frac{o_{i} w_{i,j}}{\sum_{l}^{|L|}{o_{l} w_{l,j}}}$ by dividing by the aggregated impact of all the neural weights connected to $o'_{j}$ ($\sum_{l}^{|L|}{o_{l} w_{l,j}}$).
For the influence of the output neuron connected to the given weight (2), BL computes the gradient of this output neuron to the final output; in this example of $w_{i,j}$, this gradient value is calculated as the gradient of $o'_{j}$ to the final output $O$ (i.e., $\partial O / \partial o'_{j}$). 
After quantifying these two types of influence, (1) and (2), BL multiplies them to compute the forward impact of a neural weight at last; thus, for $w_{i,j}$ in \cref{fig:target_layer}, the forward impact is $\frac{o_{i} w_{i,j}}{\sum_{l}^{|L|}{o_{l} w_{l,j}}} \frac{\partial O}{\partial o'_{j}}$. 

During the repair, we want to avoid unintentional disruption of the initially correct model behaviour as much as possible. Consequently, BL calls \textbf{ComputeGradientLoss} and \textbf{ComputeForwardImpact} twice, once with the negative inputs, $I_{neg}$ (Line 4,7) and once with the positive inputs, $I_{pos}$ (Line 5,8); for $I_{pos}$, instead of using all of them, we randomly sample the same number of positive inputs as the negative ones since a fully trained model often have far more positive inputs than the negative ones (Line 2). 
As shown respectively in Line 6 and 8, the gradient loss and the forward impact obtained with the negative inputs are divided by the corresponding values computed with the positive inputs\footnote{To avoid division-by-zero, one is added to the denominator.}. This ratio form aims to select neural weights related more to the negative inputs and less to the positive inputs. 

The result of localisation is the set of weights that constitutes the Pareto front for the combined gradient loss and the forward impact of the positive and negative inputs (Line 12): that is, the set of weights such that there is no other weight that has both a greater gradient loss (ratio) and greater forward impact (ratio) simultaneously. 
}

\small
\begin{algorithm}[!t]
\SetAlgoLined
\SetKwInOut{Input}{input}\SetKwInOut{Output}{output}
\Input{A DNN model to be repaired, \model, the set neural weights within the DNN, \textit{W}, a set of inputs that reveal the fault, $I_{neg}$, a set of inputs that correctly classified, $I_{pos}$, a loss function, \loss, 
and the number of neural weight candidates to choose based on gradient loss, $N_g$}
\Output{a set of neural weights to target for repair, $W_{t}$}

$pool \leftarrow$ $[]$\;
$I_{pos} \leftarrow RandomSample(I_{pos}, |I_{neg}|)$; 

\For{$weight$ in $W$}{ 
    $grad\_loss_{neg} \leftarrow$ |ComputeGradientLoss$(weight,\model,I_{neg},\loss)|$
    \;
    $grad\_loss_{pos} \leftarrow$ |ComputeGradientLoss$(weight,\model,I_{pos},\loss)|$\;
    $grad\_loss \leftarrow \frac{grad\_loss_{neg}}{1 + grad\_loss_{pos}}$

    $fwd\_imp_{neg} \leftarrow$ ComputeForwardImpact$(weight, \model, I_{neg})$\;
    $fwd\_imp_{pos} \leftarrow$ ComputeForwardImpact$(weight, \model, I_{pos})$\;
    $fwd\_imp \leftarrow \frac{fwd\_imp_{neg}}{1 + fwd\_imp_{pos}}$

    Add tuple $(weight, grad\_loss, fwd\_imp)$ to $pool$\;
}

$W_{t} \leftarrow $ExtractParetoFront$(pool)$\;
\Return $W_{t}$
\caption{Bidirectional Localisation of Neural Weights}\label{alg:loc}
\end{algorithm}
\normalsize

\subsection{Patch Generation} %
\label{sub:patching_the_model_bugs}

\name uses Differential Evolution (DE) to generate patches 
that repair the misclassification of a DNN image classifier. Here, we describe 
how DE is configured for repair.

\subsubsection{Differential Evolution}
\label{sec:de}

DE is a population-based metaheuristic optimisation 
algorithm that is known for its convergence speed and accuracy in continuous 
optimisation~\cite{Storn1997aa,Ronk2005aa,Das2011aa,Piotrowski2017aa}. 
\new{DE was initially designed with multidimensional real-valued functions in 
mind~\cite{Storn1997rt}, making it a good fit with \name whose aim is to quickly optimise a number of neural weights guided by a fitness function.}
A patch 
generated by \name is a set of new neural weights and is represented as a 
vector of dimension $N = |W_t|$, the number of neural weights returned by the
localisation phase.

DE starts with a pool of candidate solutions and guides the pool to achieve
higher fitness over multiple generations.
At generation $t$, let $X^t=(x^t_1, \ldots, x^t_N)$ be 
a target vector, i.e., a target solution under inspection, and $V$, a trial 
vector that may replace $X^t$ in the next generation. 
To generate $V$, DE randomly selects three vectors from its current population: $X^t_1$, $X^t_2$, and $X^t_3$, all unique and 
different from the target vector $X^t$. %
Each element of $V$, $v_i$, is generated by adding the difference between 
$x^t_{2,i}$ and $x^t_{3,i}$ to $x^t_{1,i}$. 
\cref{eq:gen_trial} shows how a candidate 
trial vector is created. 
DE employs two parameters, $F$ and $CR$, to control the mutation rate and the 
cross-over rate, respectively.
The mutation rate parameter, $F$, which 
is randomly sampled from a given range at every generation,
controls the impact of the difference as mutation: the higher $F$ is, 
the more mutated $V$ will be. As shown in Equation~\ref{eq:cand_update}, $V$ 
replaces $X^t$ if and only if $V$ is better or at least equal to $X^t$. 
This selection strategy ensures that the population of DE improves monotonically.

\begin{align}
    v_i = x^t_{1,i} + F(x^t_{2,i} - x^t_{3,i}), i \in [1,N] \label{eq:gen_trial}
\end{align}
\begin{align}
    \label{eq:cand_update}
    \vec{X}^{t+1} = 
        \begin{cases}
            \vec{V}, & \text{if } fitnesss(\vec{V})\geq fitness(\vec{X}^t)\\
            \vec{X}^t,              & \text{otherwise}
        \end{cases}
\end{align}

Algorithm \ref{alg:pseudo_de} describes the pseudo-code of DE. Crossover rate, 
$CR$, in Line 8 to 10, controls the proportion of vector elements to be 
mutated: a lower $CR$ value implies that only a small number of elements are 
mutated together, which results in DE searching each parameter separately~\cite{
Das2011aa}. Note that DE also ensures that at least one element of a 
trial vector is mutated (Line 5 and 8). We use classical DE, defining the 
range of $F$ and the value of $CR$ in advance, instead of adjusting them 
dynamically. Configuration details of DE are in 
\cref{sub:algorithm_configuration}.

\begin{algorithm}[h]

\SetAlgoLined
\SetKwInOut{Input}{input}\SetKwInOut{Output}{output}
\Input{a mutation rate, \textbf{\textit{F}}, a cross-over rate, \textbf{\textit{CR}}, the size of population \textbf{\textit{NP}}, the number of generations, \textbf{\textit{Gen}}, and a fitness function, \textbf{fitness}}
\Output{the best solution that has been found so far, \textbf{\textit{best}}}

$pop \leftarrow$ Initialise$(NP)$

\For {$gen$ in $[1,Gen]$}{
    \For {each $X$ in pop}{
        $X_1, X_2, X_3 =$ SelectRandom$(pop)$ \textcolor{blue}{//$X_1 \neq X_2 \neq X_3 \neq X$}

        $j \leftarrow$ randomly selected between $[1,N]$ %

        \For {i in $[1,N]$}{
            $r_i \leftarrow$ SelectUniform$(0,1)$

            \If {$i = j$ or $r_i \leq CR$}{
                $v_i \leftarrow x_{1,i} + F(x_{2,i} + x_{3,i})$ 
            }
        }

        \If{\textup{fitness}($V$) $\geq$ \textup{fitness}($X$)}{
            $X \leftarrow V$
        }
    }  
}

\textit{best} $\leftarrow$ SelectBest$($\textit{pop}$)$

\Return \textit{best}
\caption{Differential Evolution}\label{alg:pseudo_de}
\end{algorithm}

\subsubsection{Initialisation}

The initial population of DE must be initialised. Since \name 
attempts to search for a patch by making adjustments to neural weights using DE, it may benefit from searching in the space near the original weights.
Consequently, \name initialises the parameters of each candidate solution \new{individually: the initial value of each parameter (i.e., a localised neural weight) is sampled from a Gaussian distribution, whose mean and standard deviation are computed using all the weights in the same layer with the parameter.}

\subsubsection{Fitness Function}

\name uses a fitness function to evaluate each candidate solution and guide 
the search in DE. As in other  G\&V techniques, the fitness function of 
\name consists of two main parts: fixing misclassifications and retaining 
correct classifications. The fitness function for a candidate solution $X$ is 
defined as follows:

\begin{align}
    \label{eq:sub_fitness}
    score(i,X) = 
        \begin{cases}
            1, & \text{if } label_X(i) = label_{GT}(i)\\
            \frac{1}{Loss(i,X)+1},              & \text{otherwise}
        \end{cases}
\end{align}

\begin{equation}
fitness (X) = \sum_{i_p \in I_{pos}} score(i_p,X) + \alpha\sum_{i_n \in I_{neg}} score(i_n,X) \label{eq:fitness}
\end{equation}%

For each input $i$, \cref{eq:sub_fitness} assigns a score 
based on whether it is correctly classified or not when a candidate solution $X$ is applied.
Here, $label_X(i)$ denotes the predicted label of an input $i$ by a model patched with $X$, and $label_{GT}(i)$ is the ground truth label of the $i$.
The score is inversely 
correlated to the loss when the input is incorrectly classified, to guide the 
search towards correct classification. However, once the input is correctly 
classified, the score is simply 1.0. This is to avoid overfitting, i.e., 
reducing the loss for correct images at the cost of ignoring classification 
results for misclassified images.
\new{The set $I_{neg}$ contains inputs that reveal the misbehaviour, 
whereas the set $I_{pos}$ contains inputs that are initially processed correctly. The overall fitness is the sum of scores of both sets of inputs;  unlike the localisation, \name uses the entire set of $I_{pos}$ to compute the fitness during the repair. }

We use the hyperparameter $\alpha$ to balance the two components. A smaller
$\alpha$ means more emphasis on preserving the correct behaviour, whereas a 
larger $\alpha$ means more emphasis on correcting the misbehaviour (shown by 
$I_{neg}$). \cref{sub:fitness_evaluation} describes the details of how input 
sets, $I_{neg}$ and $I_{pos}$, are composed, as well as how the default value 
for $\alpha$ is configured.

\section{Experimental Protocols}
\label{sec:protocols}

This section outlines the experimental protocols for the empirical study and presents the research questions.

\subsection{Faults for DNNs}
\label{sec:faults}

This paper uses the term \emph{repair} to refer to the process of correcting 
the misbehaviour of a DNN. %
The terminology for the cause of the misbehaviour can, however, lead to a
philosophical question: is it possible to state that a DNN contains a 
\emph{fault} when it misbehaves? While certain types of faults in DNN models 
are explicitly committed just as in code (e.g., an incorrect choice of 
layer~\cite{Humbatova2020kt}), the kind of DNN misbehaviour that we target is 
actually anticipated due to the nature of machine learning. 
This difference may have a nontrivial impact on future work on DNN testing and 
repair. 
Note that the use of the term fault and faulty input in this paper is 
for the sake of convenience, and is not intended to answer the question about 
the nature of faults in DNNs. 

One practical issue with the type of ``faults'' we target is that it may not 
be easy to curate, document, and create a benchmark of them, as they are model 
specific and not explicitly \emph{committed}. 
To circumvent the lack of fault benchmark, we evaluate \name with two classes of faults: those artificially injected by weight perturbation, for the evaluation of the localisation phase, and those naturally emerging after full training \new{(i.e., inputs outside the training data that are \emph{not} malicious attacks generated intentionally, but nonetheless cause misbehaviour of the given model)}, for the evaluation of the patch phase. 
We solely use the artificial faults to evaluate the localization effectiveness of \name because this assessment requires the location of faults to be explicitly known. This is in contrast to patch effectiveness which is evaluated on naturally emerging faults; in this case, the exact fault location is not necessary for evaluation.

\subsection{Research Questions}
\label{sec:rqs}

We investigate the following six research questions to evaluate the effectiveness of \name. The artificial faults are only used for the evaluation of RQ1, which evaluates the localisation effectiveness of \name; meanwhile, RQs 2 to 6 use naturally emerging faults to assess the patch generation phase of \name from diverse aspects. 

\subsubsection{RQ1. Localisation Effectiveness} \emph{How effective is the 
fault localisation of \name?} 
\minorrev{To answer RQ1, we inject artificial faults into DNN models by weight perturbation. For this,} we first randomly select a neural weight among all layers whose gradient to the final output is above average. \minorrev{Here, we target only one neural weight to ensure that the observed change in model behaviour is due to the perturbation of this weight.}
\new{We then mutate the selected neural weight by adding noise from the standard normal distribution, $\mathcal{N}(0, 1)$, until it changes the behaviour of at least 0.1\% of \textit{all test inputs;} the change of behaviour includes both cases of misclassifying initial correct inputs and the opposite. \minorrev{We set this lower limit of changed model behaviour empirically. %
Since the impact of perturbing a single weight on the final classification is rather limited, the obtained limit is set at a low value, 0.1\%. We further record the actual proportion of changed behaviour, Change Ratio, to inspect the correlation between this value and the effectiveness of localisation.}
After inserting a fault into the model through the mutation, we examine if our localisation method can identify this mutated neural weight.}

\new{We compare our method against two baselines: Gradient Loss (GL) based selection, \minorrev{a variation of our fault localisation method that uses only the gradient loss values of neural weights}, %
as well as Random Selection (RS), where a random ordering is assigned to neural weights. We repeat the experiment 30 times to show that our fault localisation method works consistently.}

\subsubsection{RQ2. Patch Feasibility} \emph{Can \name \emph{repair} 
misbehaviour of a DNN model?} To show that \name can repair a DNN 
model by directly manipulating the neural weights, we randomly choose ten percent of the images that reveal misbehaviour of a DNN image classifier in \textit{the test inputs} (i.e., those unseen during the training) and investigate whether \name can repair them.
As the DE algorithm used by \name is inherently stochastic, we repeat this process 30 times.
\minorrev{This repetition includes the uniform random sampling of misbehaviour, where each sample works as a different set of misbehaviour that a model can have. Hence,} \new{we posit that any potential bias in the selection of neurons will be minimised by this repetition.} 
\minorrev{We evaluate each patch from each run on both the entire set of model misbehaviour and the sample it has seen during the patch generation, briefly checking the patch generalisability before RQ3 details it.}

Additionally, we compare the performance of \name configured with our 
localisation method, against \name using RS and GL localisation, respectively, \minorrev{
inspecting the effectiveness of different localisation methods for naturally emerging faults.}
Suppose our localisation method, BL, returns $N_{avg}$ neural weights on average. With GL
and RS, we generate an ordering as in RQ1, and select the top $N_{avg}$ weights.
To inspect whether \name can repair given misbehaviour of a DNN model, we report repair 
rate and break rate of the inputs (see details in~\cref{sub:evaluation_metric}).

\subsubsection{RQ3. Repair Generalisability} \emph{Can \name generate a patch 
that generalises to unseen data?} 

The main purpose of \name is performing focused repair on a specific type of misbehaviour. This research question verifies that \name can repair target misbehaviour, and checks whether the generated patches are also effective for unseen inputs too. 
\new{For this, we evaluate DNNs patched by \name using data unseen throughout the model training and the repair;\footnote{\cref{sub:fitness_evaluation} details how we divided the test dataset for the repair and the evaluation.}} 
We target the top 30 most frequent types of misclassifications and run \name on each misclassification type (i.e., inputs of class $A$ being misclassified as class $B$) independently. To avoid any bias in neuron selection, we include all misclassified inputs of the given type in the set of negative inputs, $I_{neg}$. Subsequently, we evaluate DNN models before and after the patch against the same types of misclassification in unseen data.

\subsubsection{RQ4. Balancing Behaviour} \emph{What is the impact of the 
balancing hyperparameter?} The hyperparameter $\alpha$ in \name balances the 
amount of focus on fixing misbehaviour and on preserving correct behaviour. We 
evaluate the impact that different values of $\alpha$ have on the repair 
outcome, using the most frequent type of misbehaviour for each dataset as an example.
Using DNN classifiers studied in RQs 1 to 3, we investigate the impact 
of varying $\alpha$. %

\subsubsection{RQ5. Comparison to State-of-the-Art\label{subsub:rq5}} \emph{How does \name 
compare to the state-of-the-art in the context of targeted repair?} 
\label{subsub:rq5_setting}
We compare \name to the recent DNN repair technique, Apricot~\cite{Zhang2019aa}. We fully\footnote{Note that subject DNNs were trained to meet a preset epoch number in Apricot~\cite{Zhang2019aa}, but our experiments found that the resulting models underfit the data under their setting.} train three DNN image classifiers using CNN architectures taken from Apricot and attempt to repair the three most frequent types of misbehaviour for each DNN classifier. 
\new{In addition to these three CNNs from Apricot, we train and repair two additional image classifiers for two different datasets that were not studied with Apricot; here, we also target the top three most frequent types of misbehaviour of each classifier.} 
While we reproduce most of the experimental settings of Apricot faithfully,
we introduce a few changes for targeted repair.
First, we modify how Apricot uses reduced Deep Learning Models (rDLMs). rDLMs are
neural networks trained on a subset of the training data, to guide repair of a DNN
based on their behaviour on individual inputs.
We modify the rDLM-based weight adjustment so that it is initiated when a specific type of misbehaviour
is discovered, instead of for every error. Second, we reduce the number of 
additional training epochs in Apricot from 20 to five, as the original setting 
resulted in inhibitive repair time. Third, we terminate Apricot when there are
no improvements for 100 batches, again for time considerations. 
We repeat repairs using \name 30 times per model, while Apricot is executed
once for each error type due to its long execution time.

\minorrev{In addition to Apricot, we compare against a \textit{retraining baseline} that takes the fault localization results of the previous step, and uses newly provided images as training data to retrain the faulty weights. In this process, we freeze other non-targeted weights and focus the learning on the weights identified by our fault localization. Negative examples are weighted with $\alpha$, equivalently to \name.}

\subsubsection{RQ6. Fairness Repair} We investigate whether \name can 
repair a more realistic and important DNN misbehaviour. Previous work has 
shown that commercial image classification APIs, trained with human faces, can 
reflect bias in the training data by showing accuracy gaps between 
genders~\cite{buolamwini2018gender}. We recreate this scenario by training a 
DNN model that classifies the gender of a given face image, using a 
pre-trained VGG architecture~\cite{Simonyan2014aa} and the Labelled Faces in 
the Wild (LFW) benchmark~\cite{LFWTech}. The classifier reflects the label 
imbalance in the dataset by showing lower class accuracy for the female 
gender. We apply \name to the trained model and see if it can successfully 
repair the label imbalance issue by patching the weights.

\subsubsection{RQ7. Model Generalisation} 

\new{To show that the use of \name is not restricted to CNN-based DNNs and further show that \name can operate on different domains than images, we evaluate \name on a completely different domain from previous RQs, namely an LSTM network trained on a text-based dataset. More specifically, we train an LSTM network model using the Twitter US Airline Sentiment dataset that contains tweets of travellers' sentiments on airlines; the goal of this model is to predict the sentiments of these tweets. We then repair the most frequent misclassification of this model using \name, which, in this case, is predicting that the tweet is neutral instead of negative. One unique characteristic of the dataset is that it provides the confidence sentiment labelers had in their label, which may help show which inputs are being handled by \name. Hence, we first evaluate the repair performance of \name, then analyse the cases in which \name fails in conjunction with the confidence of manual labellers provided by the dataset we use.}

\section{Experimental Setup}
\label{sec:experimental_setup}

This section describes the details of experimental setup.

\subsection{Subjects} %
\label{sub:subjects}

We use \new{five} well-known datasets to test the effectiveness of \name: \emph{
Fashion-MNIST}, \emph{CIFAR-10}, \emph{GTSRB}, \emph{Labeled Faces in the Wild (LFW), and \new{Twitter US Airline Sentiment}}.  

\subsubsection{Fashion-MNIST (FM)}\label{subsub:fm} 

FM~\cite{Xiao2017jy} has been introduced to overcome the shortcomings of the 
widely studied image classification benchmark, MNIST~\cite{LeCun2010vg}.
Instead of hand-written digits, FM contains 60,000 training images and 
10,000 test images of various fashion items. Each image is a 
grey-scale image of size (28,28), associated with one of ten labels. 
For this dataset, we train a neural network composed of one fully connected 
layer with 100 neurons, followed by a softmax layer with the cross-entropy loss function. We use this neural network as the DNN image classifier for the FM benchmark in RQs 1 to 4.
\new{Meanwhile in RQ5, we use the \texttt{convnet} network benchmark provided by the official FashionMNIST repository, which consists of two convolutional layers and two fully-connected layers, to evaluate \name under a more realistic scenario (in RQ5).}

\subsubsection{CIFAR-10 (C10)}\label{subsub:c10}

CIFAR-10 contains 50,000 training images and 10,000 test images with ten different classes~\cite{Krizhevsky2009kt}; each image is an RGB image of size (32,32).
For this dataset, we train a neural network with one convolutional layer with 16 filters, followed by a fully connected layer with 512 neurons and a softmax layer with cross-entropy as its loss function.
In addition to this basic neural network\new{, which we use to address RQs 1 to 4,} we also employ three DNN architectures from Apricot, namely CNN1, 2, and 3~\cite{Zhang2019aa}, \new{in RQ5.}

\subsubsection{German Traffic Sign Recognition Benchmark (GTSRB)}
\new{The GTSRB benchmark contains 51,839 images (39,209 for training, 12,630 for test) of real-world German traffic signs labelled with 43 different classes; the size of each image varies from (15,15) to (250,250). 
For this dataset, we re-implement the CNN model proposed by \textit{Team IDSIA} in the competition held at IJCNN 2011~\cite{Stallkamp2012hc}; we use this model to answer RQ5. 
Following how \textit{Team IDSIA} handled the dataset, we resize each image to be (48,48) while using only the image within the bounding box that locates the traffic sign in the centre. 
For RQs 1 to 4, we use a relatively simple model: we train a neural network composed of one convolutional layer with 16 filters, followed by a fully-connected layer with 512 neurons and a softmax layer with a cross-entropy loss function.}

\subsubsection{Labelled Faces in the Wild (LFW)}\label{subsub:lfw}

The LFW dataset~\cite{LFWTech} contains 13,233 portrait images of various 
people, each labelled with the person's identity (the original purpose of the LFW 
benchmark is to train face verification models). There are manually verified 
LFW gender labels made by Afifi \& Abdelhamed~\cite{afifi2019afif4}, which we 
use to train a classifier that predicts the gender of an image. 
The labels show the dataset has about a 10:3 ratio between male and female faces: the bias towards male faces results in an accuracy gap between the two genders, which is similar to reported cases in commercial face recognition 
APIs~\cite{buolamwini2018gender}. Consequently, we believe the trained gender 
classification model exhibits a realistic fairness issue to evaluate \name on. 
We divide this dataset into 90\% training set and 10\% test set for experiments. %

\subsubsection{Twitter US Airline Sentiment}\label{subsub:usairline}

\new{The Twitter US Airline Sentiment dataset is a collection of 14,640 real-world tweets regarding the six major US airlines; each input is in the \textit{text} format, unlike previous datasets in which the inputs were images~\cite{us_airline:tweets}. The data was gathered in February 2015, and each tweet was manually classified into three sentiment labels: positive, negative, or neutral. Along with sentiment labels, the human inspectors provided a `confidence score', which is denoted as `airline sentiment confidence', indicating how confident they were with their classification.
The tweets in this dataset vary in length. Hence, we embed these tweets using GloVe word embedding vectors, pre-trained on 6 billion tokens, with 100 dimensions for each word vector~\cite{pennington2014glove}. We drop the name of airline during this preprocessing in order to avoid bias caused by mentioning a specific airline. We split this dataset, using 80\% (11712) for the training and 20\% (2928) for the repair and evaluation (10\% for the holdout and 10\% for the evaluation).}

\subsection{Algorithm Configuration} %
\label{sub:algorithm_configuration}

DE in \name requires the $F$ and $CR$ parameters to be set in advance, as well as 
the population size and the number of generations. 
We use the default parameters of the DE implementation in \textit{scipy}~\cite{2020SciPy-NMeth}: $CR$ is set to 0.7, and $F$ is randomly sampled from (0.5, 1.0) for each generation. 
Following Piotrowski's suggestion~\cite{Piotrowski2017aa} 
that a population size of 100 suffices for problems with less than 30 
dimensions, we set the population to 100, as the number of localised neural 
weights was mostly below 30. We set the maximum number of generations to 100, but 
terminate early if the best candidate solution does not change for ten 
consecutive generations. %

\subsection{Fitness Evaluation} %
\label{sub:fitness_evaluation}

\name requires both positive and negative input sets (i.e., $I_{pos}$ and $I_{neg}$) to compute the fitness of candidate patches, as shown in \cref{eq:sub_fitness}. 
\new{In order to collect these two input sets, we divide the test set into halves, holding one half to collect the positive and negative inputs for simulating on-the-fly repair after the model deployment. We refer to this holdout input set as the \textit{validation set}. 
We then evaluate the patches generated with these inputs using the other half of the test set, referred to as the \textit{evaluation set}. 
While all RQs that involves the patch generation follow this dataset strategy, RQ2, whose aim is to inspect whether \name \textit{can} repair given misclassified inputs, is an exception; here, we collect both positive and negative inputs from the entire test set instead of from only the validation set.}
For hyperparameter $\alpha$, we set it to 10 as default, similar to previous work on Automatic Program Repair~\cite{LeClair2012}, but study the impact of 
varying its value in RQ4. \new{For RQ6 and RQ7, we set it as 2 and 1, respectively; these values are obtained empirically.}\footnote{\new{We suspect that smaller $\alpha$ worked better for RQ6 and RQ7 due to the number of unique labels in the dataset. Unlike FM, CIFAR-10, and GTSRB with either 10 or 43 labels, LFW (RQ6) and US Airline (RQ7) have only two and three labels and thereby have a higher risk of overfitting when using a larger $\alpha$ value.}}

\subsection{Model Training} %
\label{sub:model_training}

\new{We train DNN models until they do not improve after 100 consecutive epochs of training: among all the snapshots taken after each epoch, we choose the one with the highest test accuracy. This is done to prevent using under- or overfitted models. RQs 1 to 4 intend to validate the capability of \name in generating focused repair or `hot fixes' of a certain type of misbehaviour, before evaluating its actual effectiveness. %
Hence, for these RQs, we train models with relatively low capacity, whose architectures are described in \cref{sub:subjects}. These models obtain a training accuracy higher than 90\%, even with their inherently limited capacity. More precisely, for FM, the training accuracy is 95.69\%, and the test accuracy is 89.30\%; for C10, the training accuracy is 90.74\%, while the test accuracy is 72.49\%. Lastly, for GTSRB, the training and test accuracy values are 99.99\% and 97.41\%.}

We use Apricot~\cite{Zhang2019aa} as the baseline for RQ5. Apricot uses the
CIFAR-10 dataset. When comparing with Apricot, we use DNN architectures
presented in their paper (CNN1, 2, 3) for a fair comparison. The 
training process is the same as the above. Each network achieves a training accuracy of 94.27\%, 100\%, and 99.56\% and a test accuracy of 76.30\%, 83.36\%, and 76.38\%, respectively. 
\new{In addition to these three CNN models for CIFAR-10, we adopt two additional models, one for Fashion-MNIST and one for GTSRB, %
as explained in \cref{sub:subjects}. 
These two models inherently have more capacity to learn compared to those used in RQs 1 to 4. We train these models in the same way as above: the train and test accuracy are 99.98\% and 92.67\% for FM, respectively, and for GTSRB, the accuracy is 100\% for the training data and is 97.34\% for the test data.}

For RQ6, which is about fairness repair, we use a pre-trained VGG model as the starting point and tune the model for classifying face images in the LFW dataset. The training process is, again, the same as the above.  
The training accuracy of the tuned model is 99.67\% overall and 99.90\% and 98.89\% for male and female images, respectively. 
For the test dataset, the overall accuracy is 96.15\%; the accuracy for male images is 98.24\%, and for the female images, it is 86.86\%.

\new{For RQ7, which addresses a different model that is not an image classifier, we train an LSTM-based sentiment classification model; this model consists of one LSTM layer with 256 units followed by a softmax layer with the cross-entropy as the loss. %
Following the same training strategy, we obtain the training accuracy of 91.50\% and the test accuracy of 76.09\% for this model.}

\subsection{Evaluation metric}
\label{sub:evaluation_metric}

We evaluate the localisation phase of \name by computing ROC-AUC of the 
localisation methods. Each localisation method returns an ordering
of weights in terms of high suspiciousness.
Subsequently, we plot the receiver operating characteristic (ROC) curve
and calculate the area under the curve (AUC).

We evaluate the patch phase of \name from two perspectives: how successful 
patches are, and what their cost is. First, let us use the notation 
$c_{original} \rightarrow c_{predicted}$ to denote an input of class 
$c_{original}$ being misclassified as belonging to class $c_{predicted}$.
For each error type $a \rightarrow b$, we evaluate the patch effectiveness 
using a set of misclassified inputs, $I_{neg} = I_{d, (a \rightarrow b)}$, in 
which $d$ denotes the dataset these inputs are from \new{(i.e., $d \in \{\text{validation (val), evaluation (eval)}\}$).}
Consequently, $I_{d, (a \rightarrow b)} = \{i \in I_{d} | label_{GT}(i) = a \wedge label(i) = b\}$, where $label_{GT}(i)$ and $label(i)$ refer to the true and the predicted label of an input $i$.

To evaluate the degree of success \name repairs have, we define Repair Rate ($RR$) as shown in \cref{eq:repair_rate}. 
$RR$ measures how many misclassifications \name corrected.

\begin{equation}
RR(\new{I_{neg}}) = \frac{[\mbox{\# of patched input from }\new{I_{neg}}]}{|\new{I_{neg}}|}
\label{eq:repair_rate}
\end{equation}

In this equation, $I_{neg}$ is a set of inputs initially misclassified by a 
model under repair\new{; more specifically, $I_{neg}$ is a set of misclassified inputs that \name aims to repair.}
We specify two specific repair rate metrics to measure both the degree of optimisation and patch generalisation: \new{$RR_{val}$ and $RR_{eval}$. $RR_{val}=RR(I_{neg})$ measures how well \name changed the classification results of images seen during repair.}
To measure generalisability, we need a few more definitions. Suppose we aim to 
reduce instances of $a \rightarrow b$ in a trained model. We apply \name to 
the model using $I_{val, (a\rightarrow b)}$ as the negative input set, 
$I_{neg}$. However, to evaluate how general the generated patch is, we 
calculate $RR_{eval}=RR(I_{eval, (a \rightarrow b)})$, which  
captures how many unseen misclassifications of type $a \rightarrow b$ 
\name has patched.

When we measure the cost of \name, we take two things into account: the side effects and efficiency. 
To measure the side effects of \name, we use two metrics: Break Rate ($BR$) 
defined in ~\cref{eq:break_rate}, and accuracy per class. $BR$ measures how 
many initially correct inputs were broken by the repair of \name, where $I_{pos}$
is a set of inputs correctly classified by the original model.

\begin{equation}
BR = \frac{[\mbox{\# of broken input from }\new{I_{pos}}]}{|\new{I_{pos}}|}
\label{eq:break_rate}
\end{equation}

Accuracy/error per class breaks down the side effects by each class to provide a 
clearer view on how the model is affected. 
We use these two metrics flexibly between RQs, depending on their purposes.  
For efficiency, we simply measure the time it takes to repair the DNN.

\subsection{Implementation \& Environment} %
\label{sub:implementation_&_environment}

\name is implemented in Python version 3.6; our implementation of DE is 
extended from an example provided by DEAP~\cite{Fortin:2012aa}. DNN models, 
as well our baseline, are implemented using TensorFlow version 
1.12.0 or PyTorch version 1.5.1. Code and model data are publicly available from \url{https://anonymous.4open.science/r/08bf3a0a-3d52-4f61-9c42-0aa2f9c94262/}. 
All time was measured on machines equipped with Intel Core i7 
CPU, 32GB RAM, and NVidia TitanX GPU.

\section{Results and Analysis}
\label{sec:results}

This section reports and discusses the results of the empirical evaluation, 
and answers the research questions described in \cref{sec:rqs}.

\subsection{RQ1: Localisation Effectiveness}

\begin{figure*}[t!]
    \begin{subfigure}{0.32\textwidth}
        \includegraphics[width=\textwidth, trim=0mm 0mm 0mm 0mm, clip]{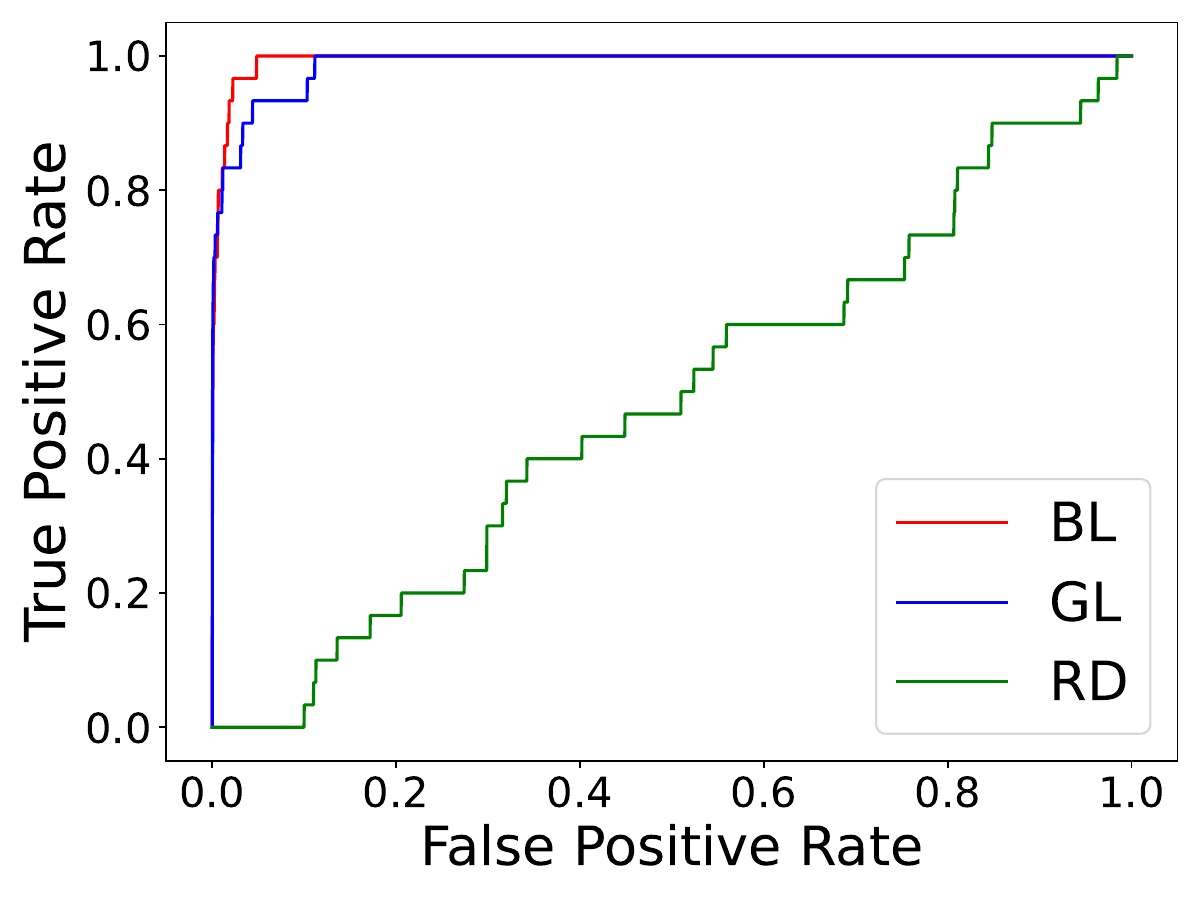}
        \caption{CIFAR-10}
        \label{fig:rq1_C10}
    \end{subfigure}
    \begin{subfigure}{0.32\textwidth}
        \includegraphics[width=\textwidth, trim=0mm 0mm 0mm 0mm, clip]{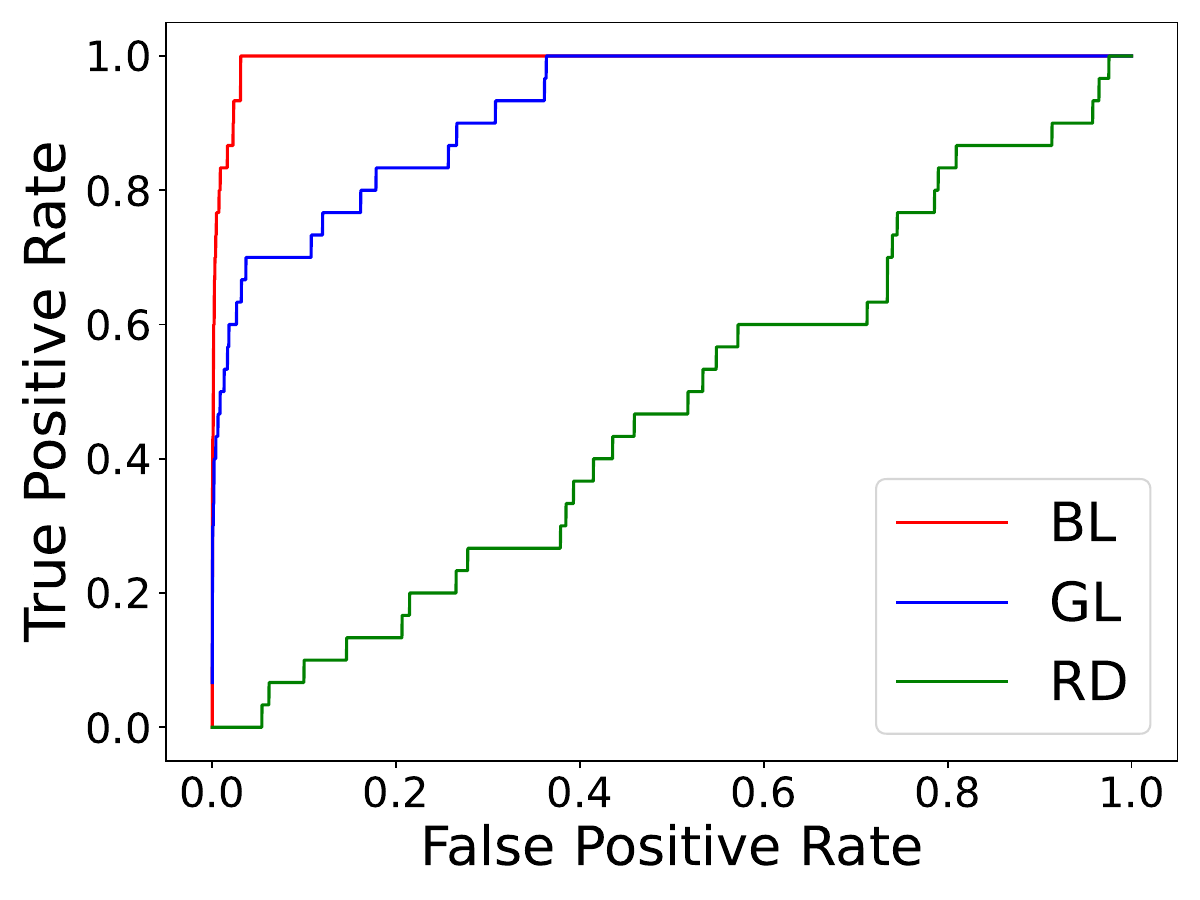}
        \caption{Fashion MNIST}
        \label{fig:rq1_fm}
    \end{subfigure}
    \begin{subfigure}{0.32\textwidth}
        \includegraphics[width=\textwidth, trim=0mm 0mm 0mm 0mm, clip]{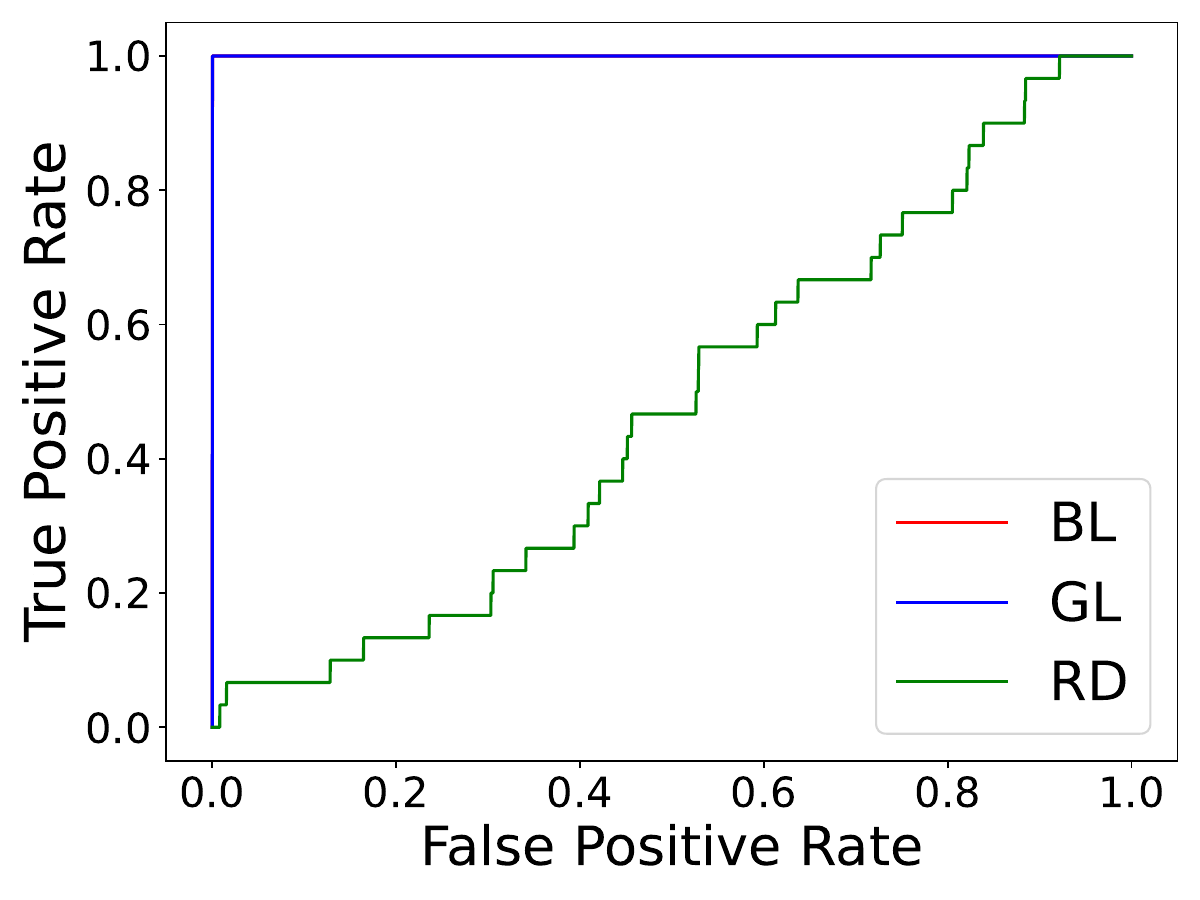}
        \caption{GTSRB}
        \label{fig:rq1_gtsrb}
    \end{subfigure}

    \caption{
    ROC curves for each localisation method. 
    The lines indicate the median value over 30 runs; 
    the shades represent the true positive value range including 90\% of all runs. }
    \label{fig:rq1_roc}
\end{figure*}

\new{\cref{tab:rq1_fl} reports the ratio of inputs whose behaviour was changed by a mutated neural weight, namely, a fault, and the ROC-AUC values of different weight prioritisation schemes including ours. 
Figure~\ref{fig:rq1_roc} visualises the average ROC curves from 30 runs.} %

\new{As shown in the Change Ratio column in \cref{tab:rq1_fl}, among the three data sets, Fashion-MNIST (FM) is the least affected by the perturbation of single neural weight, while GTSRB is most affected. Both the proposed localisation method, BL, and the baseline GL outperform RS in all three datasets, assigning the mutated neural weight a higher inspection priority. 
Compared to GL, BL obtains better performance in FM and C10, and performs similarly in GTSRB. The difference between the localisation performance is the most pronounced in FM, followed by C10 and GTSRB.}

\new{We suspect these variations in the performance difference are related to the impact of a single weight mutation on the model, which the Change Ratio reflects. 
As described in RQ1 in \cref{sec:rqs}, we did not set an upper limit on the impact of a mutated neural weight on initial model behaviour: the perturbed neural weight can change more than 0.1\% of the inputs. 
Indeed, the average change ratio varies from 0.61\% to 4.94\%, depending on the dataset. GL prioritises neural weights in descending order of their absolute gradient loss values. Hence, the more a single neural weight alone can affect the model behaviour, the greater its absolute gradient loss value is likely to be.
As a result, compared to FM, GL obtains comparable performance to our method BL in C10 and GTSRB, the two datasets with a relatively higher change ratio.}
\minorrev{In GTSRB, both BL and GL obtain a close-to-perfect performance, obtaining 1.0 for rounded ROC-AUC. We conjecture that this is due to the GTSRB dataset both containing more labels and fewer training data points. Together, these characteristics make the trained DNN model more sensitive toward small changes, especially when it obtains high accuracy. Indeed, the GTSRB model has a relatively higher change ratio and accuracy than the models of FM and C10.}

\new{We designed this localisation experiment to have a single faulty neural weight. %
However, in practice, multiple neural weights jointly play roles in the model misbehaviour, and therefore, the effect of a single neural will be less dramatic than in this experiment. 
Consequently, the notable performance of the proposed method, BL, in FM compared to other baselines implies that BL can effectively discriminate neural weights that cause the behavioural changes in the model.} \\

\noindent\fbox{\parbox{\textwidth}{%
\textbf{Answer to RQ1: } From the results, we conclude that the proposed BL 
method can effectively identify neural weights that are responsible for 
misbehaviour.}}

\begin{table}[ht]
    \centering
    \caption{The ratio of changed inputs (Change Ratio) and fault localisation accuracy (ROC-AUC) of the proposed method (BL), Gradient Loss (GL), and Random Selection (RS)
    \label{tab:rq1_fl}}
    \scalebox{0.9}{
    \begin{tabular}{l|ccc|c|c|c}
    \toprule
        \multirow{2}{*}{Subject} & \multicolumn{3}{c|}{Change Ratio (\%)} & \multirow{2}{*}{\textit{BL}} & \multirow{2}{*}{\textit{GL}} & \multirow{2}{*}{\textit{RS}} \\
        &  avg & min & max & & & \\
    \midrule

    C10 & 0.61 & 0.12 & 2.43 & \Numpfour{0.9947}    & \Numpfour{0.988} & \Numpfour{0.481} \\%
    FM & 0.17 & 0.1 & 0.42 & \Numpfour{0.9945} & \Numpfour{0.9237}    & \Numpfour{0.4726} \\
    GTSRB & 4.94 & 4.21 & 6.05 & \Numpfour{0.99995288} & \Numpfour{0.99998642}    & \Numpfour{0.47744537} \\

    \bottomrule
    \end{tabular}
    }
\end{table} 

\subsection{RQ2: Patch Feasibility} %
\label{sub:rq2}

Table~\ref{tab:rq2} contains the average repair rate ($RR_{val} = RR(I_{neg})$) \new{ computed for the randomly selected ten percent of initial misclassified inputs,} and the break rate ($BR$) from 30 repeated runs of \name with three different localisation methods; \minorrev{for the sake of simplicity, we denote $RR_{val}$ as $RR$ in Table~\ref{tab:rq2}.}
\new{To summarise the overall repair cost, we further report the ratio of the post-patch model accuracy to the initial accuracy, $R_{acc}$, and %
\minorrev{the difference between the number of correct inputs before and after the patch, \#$_{\text{diff}}$}, which is negative for performance degradation and positive for improvement,} \minorrev{in the third column for each localisation method.}

\new{In order to mitigate potential bias from the input selection, at each run, we randomly select 10\% of the total misclassified test inputs for \name to repair: 277, 107 and 125 inputs are selected for C10, FM, and GTSRB.
Since the average number of localised neural weights is 11.6, 7.6, and 14.3 respectively for C10, FM, and GTSRB, GL and RS localisation methods were configured to select 12, 8, and 14 weights for C10, FM, and GTSRB.}

\new{With BL, \name successfully repairs on average 10.8\% of the randomly selected misbehaviour in C10, while breaking the average 0.31\% of the initially correct inputs ($I_{pos}$). RS barely modifies the network, and GL pays much more for the repair (i.e., breaks more inputs) than BL in C10.
The $R_{acc}$ column in Table~\ref{tab:rq2} summarise the overall cost of the repair; compared to using BL for localisation, the misclassified inputs increase more than seven times in GL, as shown in the \#$_{\text{diff}}$ column (i.e., -14.4 for BL and -103.4 for GL). %
}

\new{For FM, \name with BL repairs 8.7\% of the given misclassifications and breaks 0.7\% of the initially correct ones, on average. 
Although \name repairs a little more when using GL (i.e., 9.5\%), it also breaks more correct inputs. Similarly to what we observed in C10, RS repairs and breaks less than the other two. The accuracy decreases with all the three localisation methods but only slightly, retaining more than 99\% of the initial accuracy. }

\new{For GTSRB, \name breaks less and repairs more or the same with BL compared to GL. For instance, \name repairs 13.0\% of the given misclassifications using BL and 10.7\%, a little less than BL, with GL. 
In terms of BR, BL outperforms GL by a small margin; the accuracy improves with BL and degrades with GL.
Overall, the effect of \name is smaller in degree for GTSRB than for C10 and FM. We suspect this is due to the greater number of categories in the GTSRB dataset. As GTSRB has more categories (43) than the other two (10), the selected negative inputs of GTSRB likely include more various types of misclassifications than those of C10 and FM. \name repairs the given misbehaviour (i.e., misclassifications) of the model ($I_{neg}$) by directly modifying the neural weights specifically related to them. Hence, if there are multiple types of misbehaviour in $I_{neg}$, the overall repair effort of \name becomes distributed, resulting in less effective performance.  
Here, again, RS barely makes any meaningful change. %
}

\new{\name repairs the randomly chosen 10\% of negative inputs instead of all of them in this experiment. Hence, we additionally compute the overall repair rate ($RR_{overall}$), $RR$ calculated for the entire set of negative inputs instead of the randomly chosen 10\%, to reduce the risk of drawing biased conclusions based on the selected inputs. Overall, $RR_{overall}$ shows a similar trend to that of $RR(=RR(I_{neg}))$, supporting the generalisability of our findings.\footnote{RQ3 will detail the generalisability of \name.}}

While the underperformance of RS is what we expected (i.e., repair and break less), %
the results of GL require closer analysis. Since GL localises neural weights by
following the gradient loss, we suspect that GL risks localising weights with
small forward impact. Consequently, \name needs to make more dramatic changes
to these weights in order to change the final classification outcome. In turn,
this may have a greater effect on other inputs. BL does not suffer from these issues, and we believe that it can identify locations that have maximal patch
effect with minimal modification. 
In summary, we conclude that \name can patch
misclassifications, for which BL is the most effective localisation.\\

\noindent\fbox{\parbox{\textwidth}{%
\textbf{Answer to RQ2: } 
\name can repair random misclassifications of a DNN while preserving most of 
the initially correct behaviour: \new{repairing 10.8\%, 8.7\% and 13.0\% of misclassifications in the test data, while breaking only 0.3\%, 0.7 and 0.7\% for C10, FM and GTSRB, respectively.} The results suggest that BL is the most effective among the studied approaches in terms of balancing between repairing and preserving correct behaviour.
}}

\begin{table}[h]
\centering
\caption{\minorrev{The repair performance ($RR$ and $RR_{overall}$) and cost ($BR$ and $R_{acc}$ (\#$_{\text{diff}}$)) of \name using bidirectional localisation method (BL), gradient loss (GL), and random selection (RS) over 30 runs.}\label{tab:rq2}}
\scalebox{0.79}{
\begin{tabular}{c|rrr|rrr|rrr}
\toprule%
        & \multicolumn{3}{c|}{\textit{BL}} & \multicolumn{3}{c|}{\textit{GL}} & \multicolumn{3}{c}{\textit{RS}} \\
Subject & \minorrev{$RR$ / $RR_{\text{overall}}$} & $BR$ & R$_{acc}$ (\#$_{\text{diff}}$) & \minorrev{$RR$ / $RR_{\text{overall}}$} & $BR$ & R$_{acc}$ (\#$_{\text{diff}}$) & \minorrev{$RR$ / $RR_{\text{overall}}$} & $BR$ & R$_{acc}$ (\#$_{\text{diff}}$) \\
\midrule

C10 & \Numpthree{0.1083} / \Numpthree{0.0742} & \Numpthree{0.0305} & \Numpthree{0.998} (-14.4)
& \Numpthree{0.1357} / \Numpthree{0.1169} & \Numpthree{0.0593} & \Numpthree{0.9857} (-103.4)
& \Numpthree{0.0174} / \Numpthree{0.0066} & \Numpthree{0.0013} & \Numpthree{1.0013} (9.1)\\

FM & \Numpthree{0.0866} / \Numpthree{0.0423} & \Numpthree{0.007} & \Numpthree{0.9981} (-17.2)
& \Numpthree{0.095} / \Numpthree{0.0473} & \Numpthree{0.0079} & \Numpthree{0.9978} (-20.0)
& \Numpthree{0.0542} / \Numpthree{0.0206} & \Numpthree{0.0032} & \Numpthree{0.9993} (-6.2) \\%

GTSRB & \Numpthree{0.1304} / \Numpthree{0.0692} & \Numpthree{0.007} & \Numpthree{1.0007} (7.5)
& \Numpthree{0.1067} / \Numpthree{0.0683} & \Numpthree{0.0082} & \Numpthree{0.9993} (-8.1)
& \Numpthree{0.0093} / \Numpthree{0.0039} & \Numpthree{0.0} & \Numpthree{1.0004} (4.8) \\%

\bottomrule
\end{tabular}
}
\end{table}

\subsection{RQ3: Repair Generalisability} %
\label{sub:rq3}

\begin{figure*}[t!]
    \centering
    \begin{subfigure}{0.49\textwidth}
        \includegraphics[width=\textwidth, trim=0mm 0mm 0mm 0mm, clip]{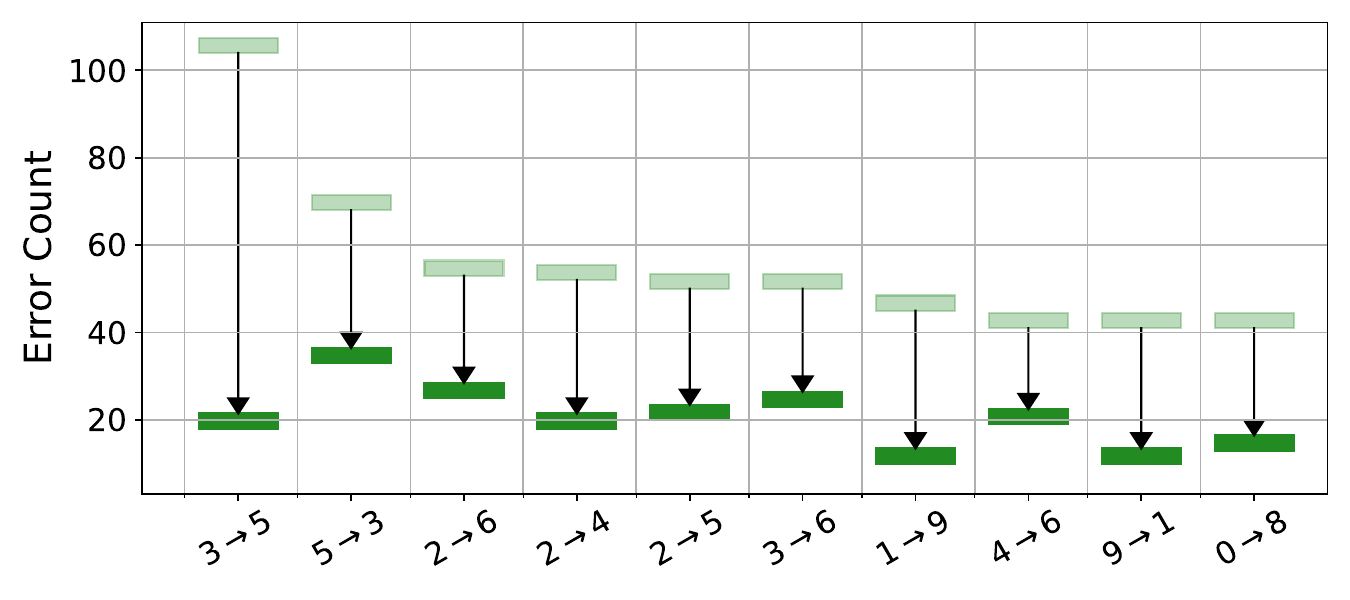}
        \caption{CIFAR-10 (validation)}
        \label{fig:rq3_cifar10_train}
    \end{subfigure}
    \begin{subfigure}{0.49\textwidth}
        \includegraphics[width=\textwidth, trim=0mm 0mm 0mm 0mm, clip]{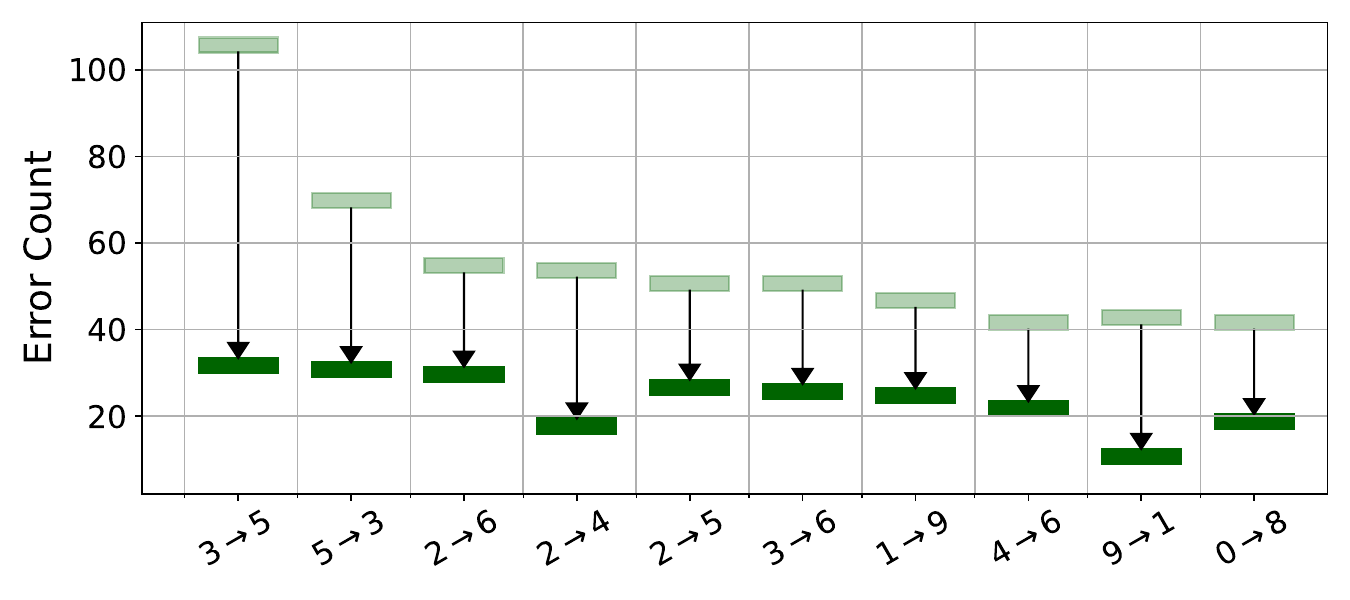}
        \caption{CIFAR-10 (evaluation)}
        \label{fig:rq3_cifar10_test}
    \end{subfigure}

    \begin{subfigure}{0.49\textwidth}
        \includegraphics[width=\textwidth, trim=0mm 0mm 0mm 0mm, clip]{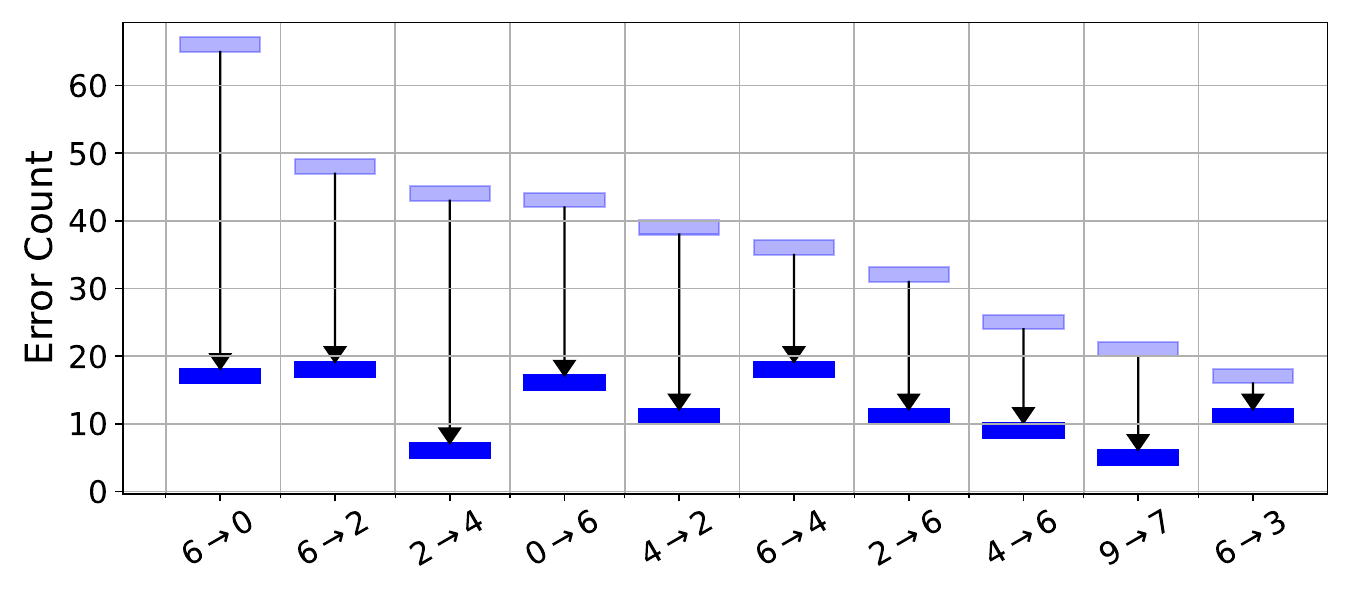}
        \caption{Fashion-MNIST (validation)}
        \label{fig:rq3_fm_train}
    \end{subfigure}
    \begin{subfigure}{0.49\textwidth}
        \includegraphics[width=\textwidth, trim=0mm 0mm 0mm 0mm, clip]{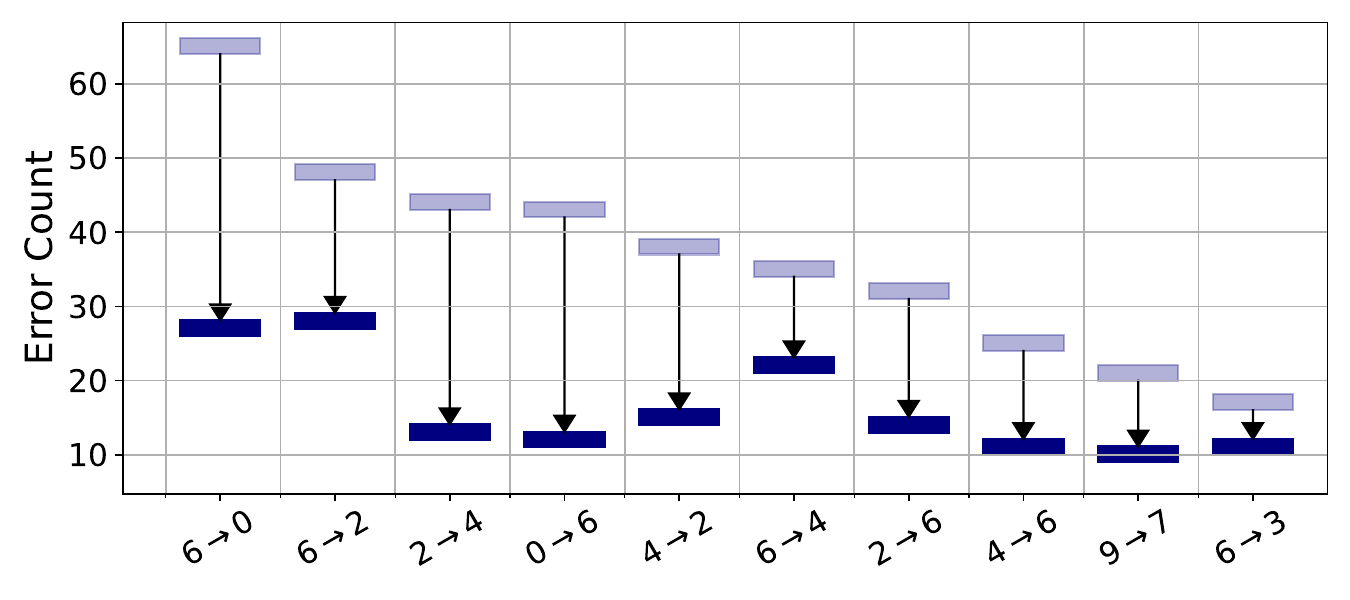}
        \caption{Fashion-MNIST (evaluation)}
        \label{fig:rq3_fm_test}
    \end{subfigure}

    \begin{subfigure}{0.49\textwidth}
        \includegraphics[width=\textwidth, trim=0mm 0mm 0mm 0mm, clip]{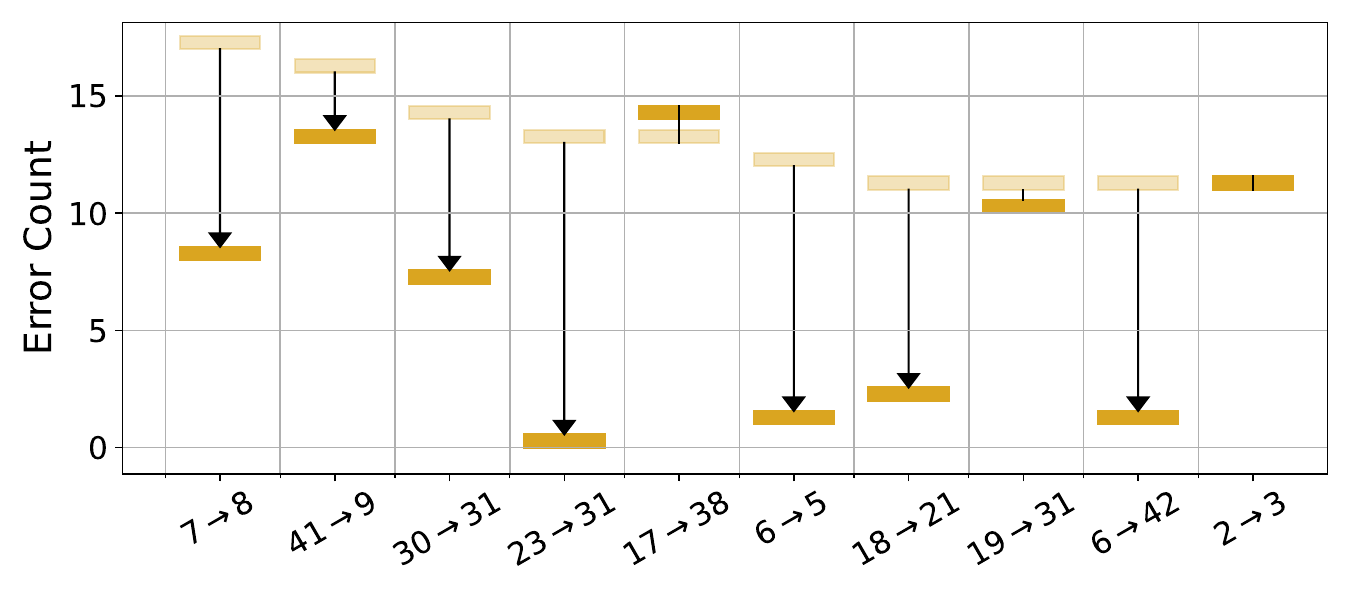}
        \caption{GTSRB (validation)}
        \label{fig:rq3_gtsbr_train}
    \end{subfigure}
    \begin{subfigure}{0.49\textwidth}
        \includegraphics[width=\textwidth, trim=0mm 0mm 0mm 0mm, clip]{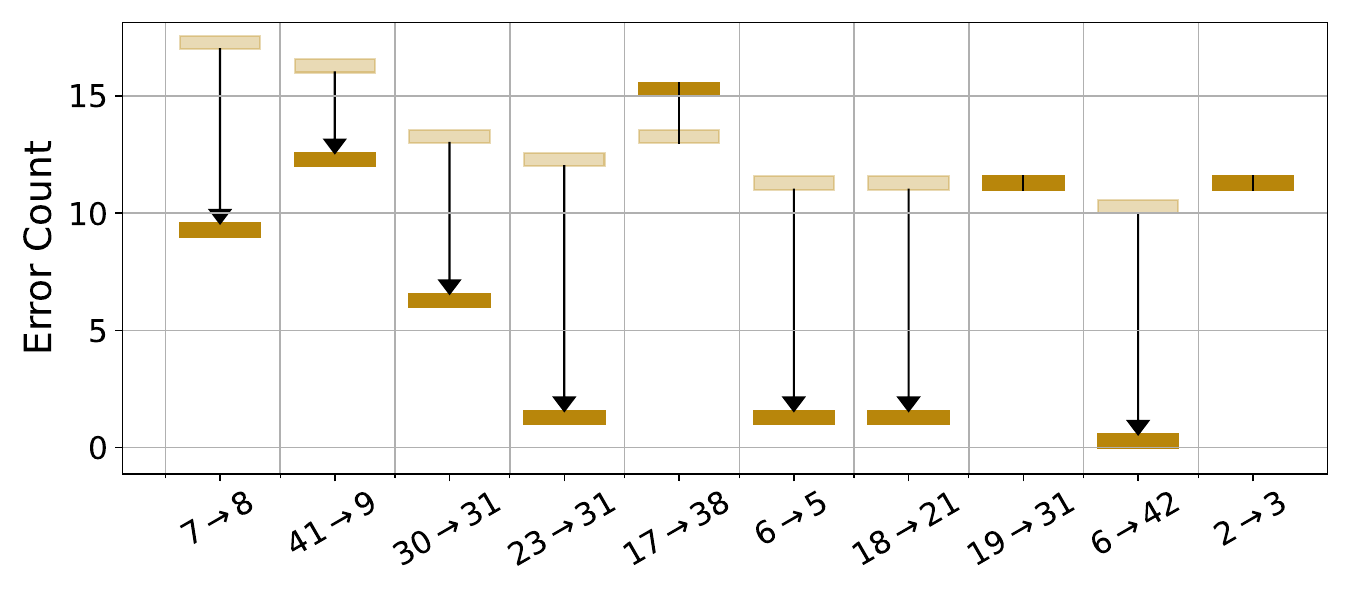}
        \caption{GTSRB (evaluation)}
        \label{fig:rq3_gtsbr_test}
    \end{subfigure}
    
    \caption{
    \new{Visualisation of repair rate for the top 10 most frequent types of misclassification. The y-axis is the number of misclassifications that belong to each type. The light and dark bars denote the number of misclassifications before and after \name, respectively. The arrow from the light bar to the dark bar depicts the decrease in the number of misclassifications.}}
    \label{fig:rq3_alpha}
\end{figure*}

In RQ3, we aim to verify the generalisability of patches made by \name to fix specific misbehaviour, by
reporting \new{both $RR_{val}$ and $RR_{eval}$}.
\cref{tab:rq3} reports the number of misclassifications repaired by \name in the \new{validation and the evaluation dataset.} The results suggest that a patch effective for a specific type of misclassifications in the validation data is also effective against the same type of misclassifications in 
the evaluation data. \new{For example, \name successfully generates patches that fix 83\% (86) of the $3 \rightarrow 5$ misclassifications in the validation data of C10 on average; the same patches can repair 71\% (74) of the same type of misclassified inputs in the evaluation data. 
Similarly, the patches that repair the average 75\% (49) of $6 \rightarrow 0$ misclassifications in FM can also repair the average 59\% (38) of the same misclassifications in the evaluation data.
In GTSRB, although the degree of repair is smaller in general, the patches that repair 53\% (9) of the most frequent misclassifications ($7 \rightarrow 0$) can also fix 47\% (8) of the same type of misclassifications in the evaluation data.
} The trend is similar in other misclassification types as well. \cref{fig:rq3_alpha} visualises the overall results of \cref{tab:rq3}.

\cref{tab:rq3_distr} shows the overall generalisability results across all 30 
studied misclassification types. We report the mean and standard deviation of the ratio between $RR_{val}$ and $RR_{eval}$, aggregated over the top 1, 10, 20, and 30 most frequent misclassification types\new{; for the top 1, we drop the standard deviation as there is only one ratio computed for the most frequent errors.} The higher this ratio is, the more the patch generalises to unseen misclassifications in the test data.
\new{For this evaluation, we excluded the cases where \name cannot repair the given misclassifications at all ($RR_{val} = 0$); the numbers of such cases within the top 1, 10, 20 and 30 types are reported sequentially in \cref{tab:rq3_distr}, next to the project identifier.} 

Regarding the top 10 most frequent misclassification types in C10, \name keeps 88.48\% of the repair rate (on average) with a standard deviation of only 0.1818 when moving from an observed dataset (validation) to an unobserved dataset (evaluation).
\new{\name shows similar results in FM, retaining 76.49\% of the repair rate in the validation data for the top 10. In GTSRB, with both the average rate and the standard deviation of this ratio often greater than one, \name shows unstable performance compared to the other two. %
As can be inferred from \cref{tab:rq3}, the number of misclassifications for each error type is generally smaller in GTSRB than in C10 and FM. This insufficient number of misclassification of GTSRB may explain the erratic performance we observed since it could increase the risk of generating close-to-random or over-fitted patches. The relatively low repair rate of \name in GTSRB in \cref{tab:rq3} further supports this suspicion. Nonetheless, \name still retains 66\% of the repair rate for the most frequent type in GTSRB, promising its generalisability.}

While this ratio tends to decrease as we gradually include less frequent misclassification types, for the top 30 most frequent types, \name successfully retains more than half of the its repair rate against unseen data for all three datasets. \\

\noindent\fbox{\parbox{\textwidth}{%
\textbf{Answer to RQ3:} 
The repair results for the 30 most frequent misclassification types show that \name can fix
targeted misclassifications, and that the patches generated by \name generalise to unseen images.
}}

\begin{table*}[ht]
\centering
\caption{\new{The number of inputs repaired by \name for the top ten most frequent misclassification types (Freq). The values in brackets are repair rates computed for each type.}}\label{tab:rq3}
\scalebox{0.73}{
\begin{tabular}{ll|rrrrrrrrrr}
\toprule
Freq. & & 1 & 2 & 3 & 4 & 5 & 6 & 7 & 8 & 9 & 10 \\
\midrule

\multirow{3}{*}{\emph{C10}} 
& fault & $3 \rightarrow 5$ & $5 \rightarrow 3$ & $2 \rightarrow 6$ & $2 \rightarrow 4$ & $2 \rightarrow 5$ & $3 \rightarrow 6$ & $1 \rightarrow 9$ & $4 \rightarrow 6$ & $9 \rightarrow 1$ & $0 \rightarrow 8$\\

& val & 86 (\Numptwo{0.8269230769230769})  & 35 (\Numptwo{0.5147058823529411})  & 28 (\Numptwo{0.5283018867924528})  & 34 (\Numptwo{0.6538461538461539})  & 30 (\Numptwo{0.6})  & 27 (\Numptwo{0.54})  & 35 (\Numptwo{0.7777777777777778})  & 22 (\Numptwo{0.5365853658536586})  & 31 (\Numptwo{0.7560975609756098})  & 28 (\Numptwo{0.6829268292682927}) \\%

& eval & 74 (\Numptwo{0.7115384615384616})  & 39 (\Numptwo{0.5735294117647058})  & 25 (\Numptwo{0.4716981132075472})  & 36 (\Numptwo{0.6923076923076923})  & 24 (\Numptwo{0.4897959183673469})  & 25 (\Numptwo{0.5102040816326531})  & 22 (\Numptwo{0.4888888888888889})  & 20 (\Numptwo{0.5})  & 32 (\Numptwo{0.7804878048780488})  & 23 (\Numptwo{0.575}) \\

\midrule
\multirow{3}{*}{\emph{FM}} 
& fault & $6 \rightarrow 0$ & $6 \rightarrow 2$ & $2 \rightarrow 4$ & $0 \rightarrow 6$ & $4 \rightarrow 2$ & $6 \rightarrow 4$ & $2 \rightarrow 6$ & $4 \rightarrow 6$ & $9 \rightarrow 7$ & $6 \rightarrow 3$\\

& val & 49 (\Numptwo{0.7538461538461538})  & 30 (\Numptwo{0.6382978723404256})  & 38 (\Numptwo{0.8837209302325582})  & 27 (\Numptwo{0.6428571428571429})  & 28 (\Numptwo{0.7368421052631579})  & 18 (\Numptwo{0.5142857142857142})  & 21 (\Numptwo{0.6774193548387096})  & 16 (\Numptwo{0.6666666666666666})  & 16 (\Numptwo{0.8})  & 6 (\Numptwo{0.375}) \\

& eval & 38 (\Numptwo{0.59375})  & 20 (\Numptwo{0.425531914893617})  & 31 (\Numptwo{0.7209302325581395})  & 31 (\Numptwo{0.7380952380952381})  & 23 (\Numptwo{0.6216216216216216})  & 13 (\Numptwo{0.38235294117647056})  & 18 (\Numptwo{0.5806451612903226})  & 14 (\Numptwo{0.5833333333333334})  & 11 (\Numptwo{0.55})  & 6 (\Numptwo{0.375}) \\

\midrule
\multirow{3}{*}{\emph{GTSRB}} 
& fault & $7 \rightarrow 8$ & $41 \rightarrow 9$ & $30 \rightarrow 31$ & $23 \rightarrow 31$ & $17 \rightarrow 38$ & $6 \rightarrow 5$ & $18 \rightarrow 21$ & $19 \rightarrow 31$ & $6 \rightarrow 42$ & $2 \rightarrow 3$\\

& val & 9 (\Numptwo{0.5294117647058824})  & 3 (\Numptwo{0.1875})  & 7 (\Numptwo{0.5})  & 13 (\Numptwo{1.0})  & -1 (\Numptwo{-0.07692307692307693})  & 11 (\Numptwo{0.9166666666666666})  & 9 (\Numptwo{0.8181818181818182})  & 1 (\Numptwo{0.09090909090909091})  & 10 (\Numptwo{0.9090909090909091})  & 0 (\Numptwo{0.0}) \\

& eval & 8 (\Numptwo{0.47058823529411764})  & 4 (\Numptwo{0.25})  & 7 (\Numptwo{0.5384615384615384})  & 11 (\Numptwo{0.9166666666666666})  & -2 (\Numptwo{-0.15384615384615385})  & 10 (\Numptwo{0.9090909090909091})  & 10 (\Numptwo{0.9090909090909091})  & 0 (\Numptwo{0.0})  & 10 (\Numptwo{1.0})  & 0 (\Numptwo{0.0}) \\

\bottomrule
\end{tabular}}
\end{table*}

\begin{table}[ht]
\centering
\caption{\new{The ratio between $RR_{used}$ and $RR_{eval}$, computed for the top \emph{N} frequent types of misclassification ($N \in\{1,10,20,30\}$).
The four numbers next to the project are the number of cases where \name completely fails to repair the given type of misclassifications for the top 1, 10, 20, and 30 frequent misclassification types. 
}
}\label{tab:rq3_distr}
\scalebox{0.8}{
\begin{tabular}{l|r|rr|rr|rr}
\toprule
top $N$ & 1 & \multicolumn{2}{c|}{10} & \multicolumn{2}{c|}{20} & \multicolumn{2}{c}{30} \\
&  & mean & std & mean & std & mean & std \\
\midrule

\emph{C10} (0/0/0/0) & 0.8571 & 0.8848 & 0.1818 & 0.7872 & 0.2850 & 0.7865 & 0.2935\\
\emph{FM} (0/0/1/3) & 0.7876 & 0.7649 & 0.2093 & 0.8062 & 0.2786 & 0.7058 & 0.3771\\
\emph{GTSRB} (0/2/2/5) & 0.6667 & 1.3735 & 1.6264 & 1.0188 & 1.1531 & 0.8282 & 1.0507\\

\bottomrule
\end{tabular}}
\end{table}  

\subsection{RQ4: Balancing Behaviour}
\label{sub:rq4}

\begin{figure*}[ht!]
    \centering
    \begin{subfigure}{0.4\textwidth}
        \includegraphics[width=\textwidth, trim=0mm 5mm 0mm 0mm, clip]{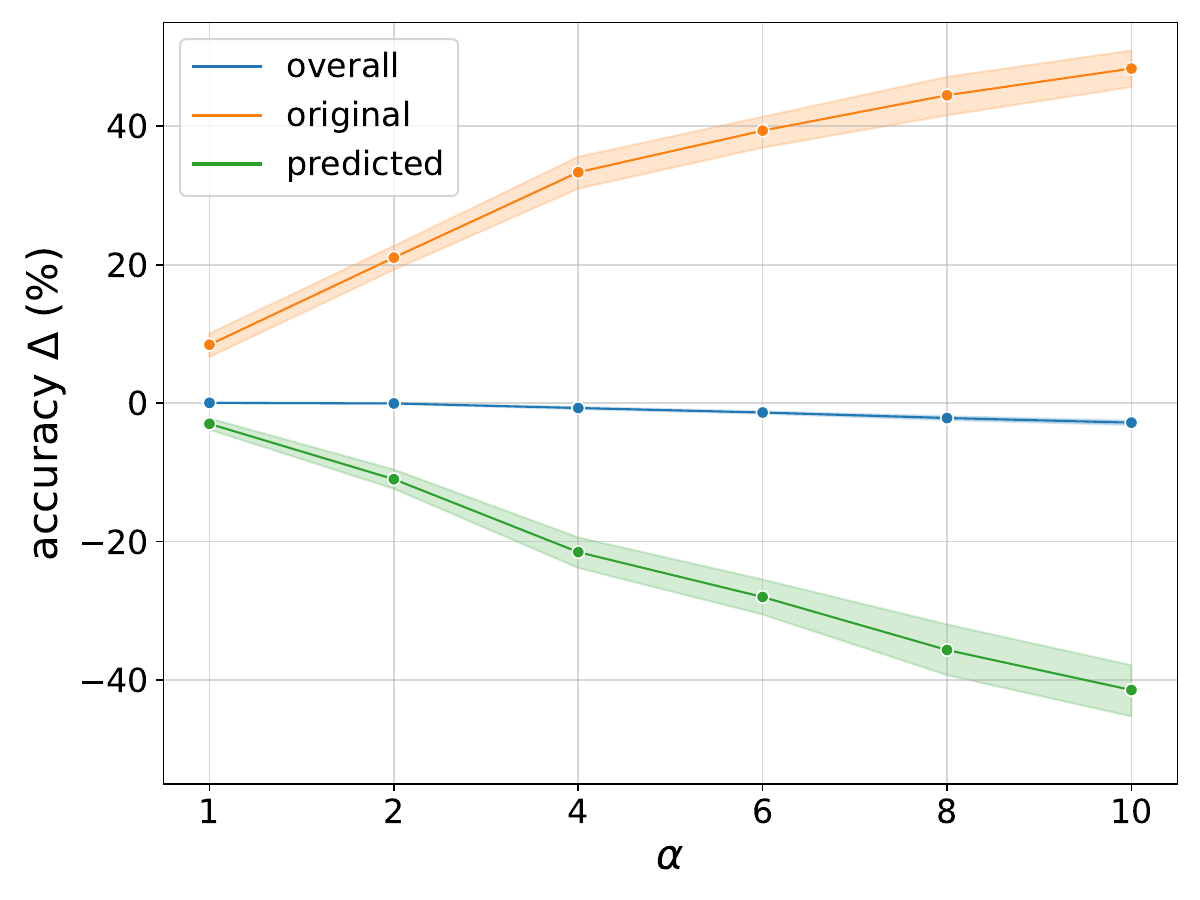}
        \caption{CIFAR-10 (validation)}
        \label{fig:rq4_cifar10_used}
    \end{subfigure}
    \begin{subfigure}{0.4\textwidth}
        \includegraphics[width=\textwidth, trim=0mm 5mm 0mm 0mm, clip]{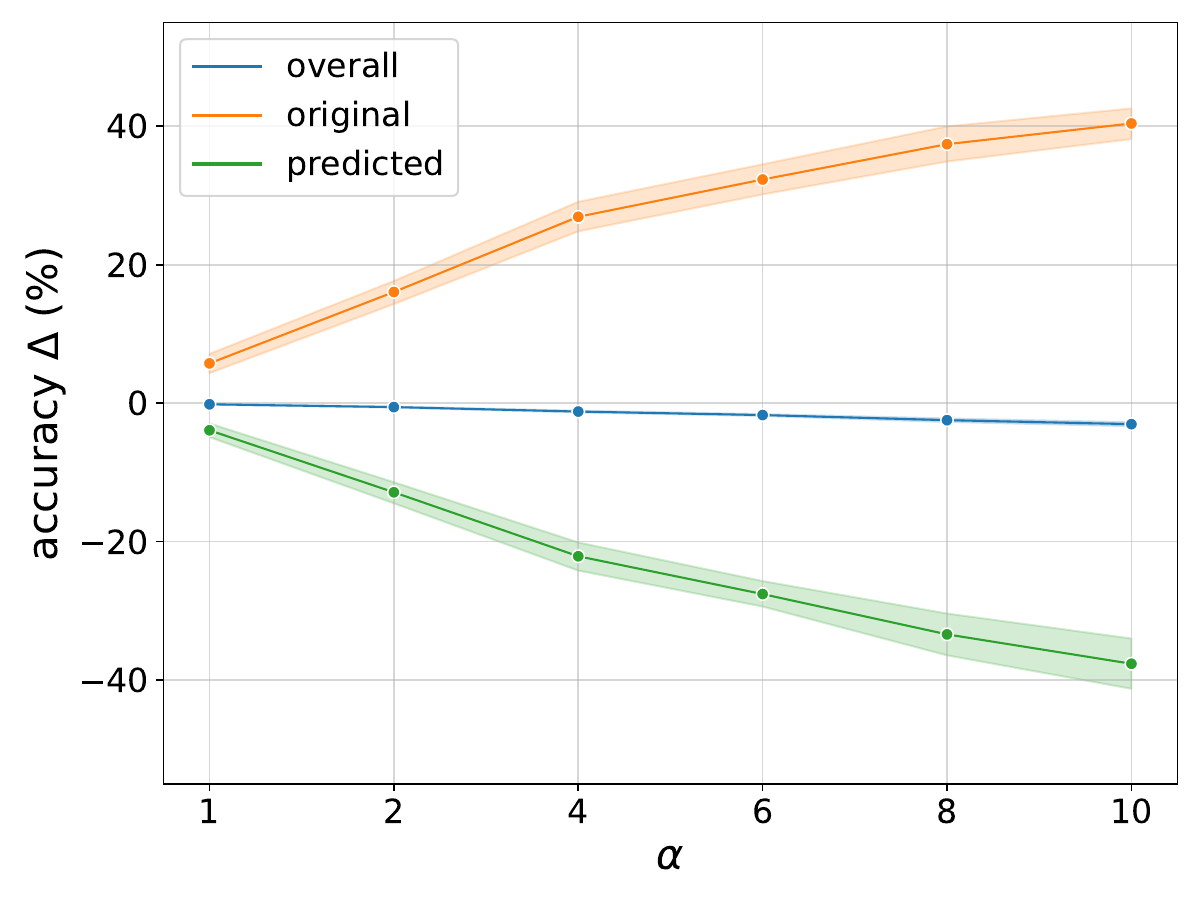}
        \caption{CIFAR-10 (evaluation)}
        \label{fig:rq4_cifar10_eval}
    \end{subfigure}
    \vspace{0.5em}

    \begin{subfigure}{0.4\textwidth}
        \includegraphics[width=\textwidth, trim=0mm 5mm 0mm 0mm, clip]{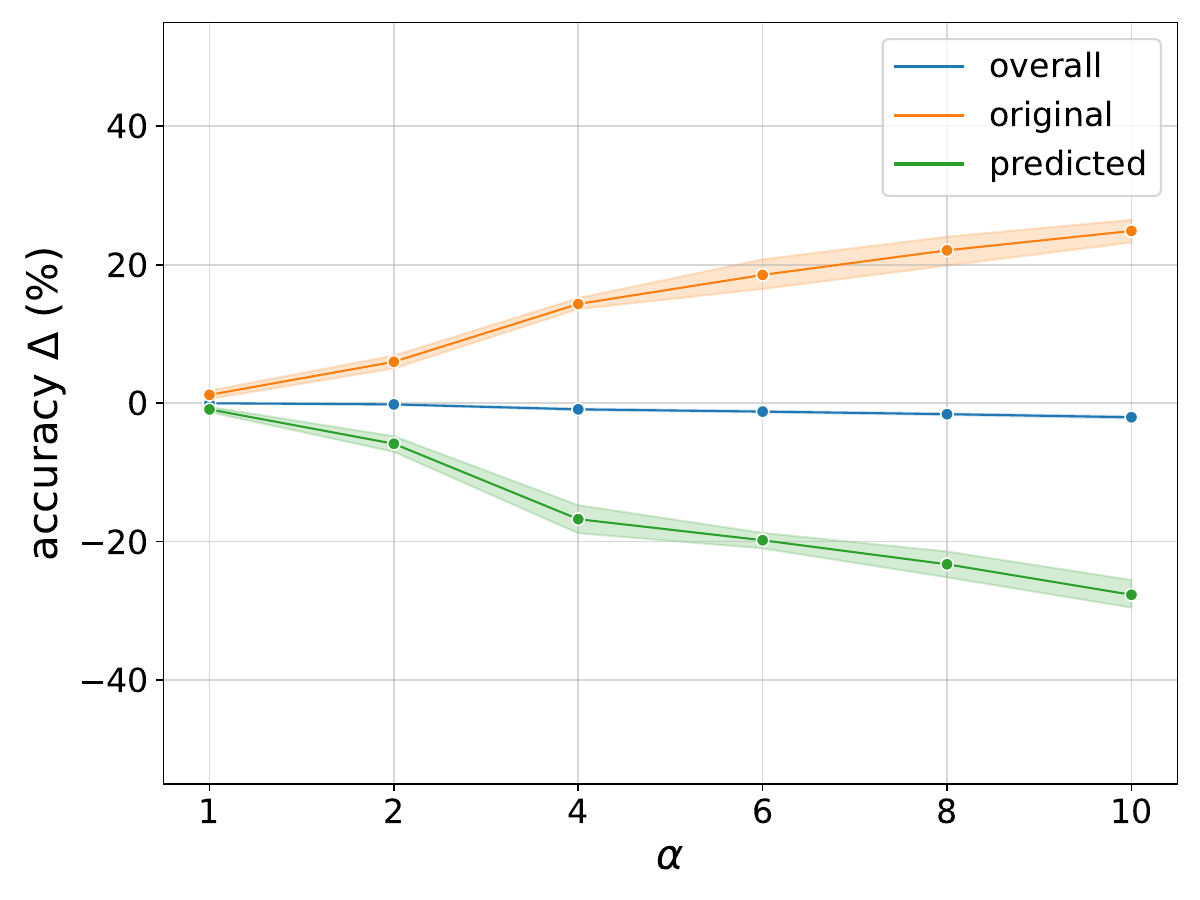}
        \caption{Fashion-MNIST (validation)}
        \label{fig:rq4_fm_used}
    \end{subfigure}
    \begin{subfigure}{0.4\textwidth}
        \includegraphics[width=\textwidth, trim=0mm 5mm 0mm 0mm, clip]{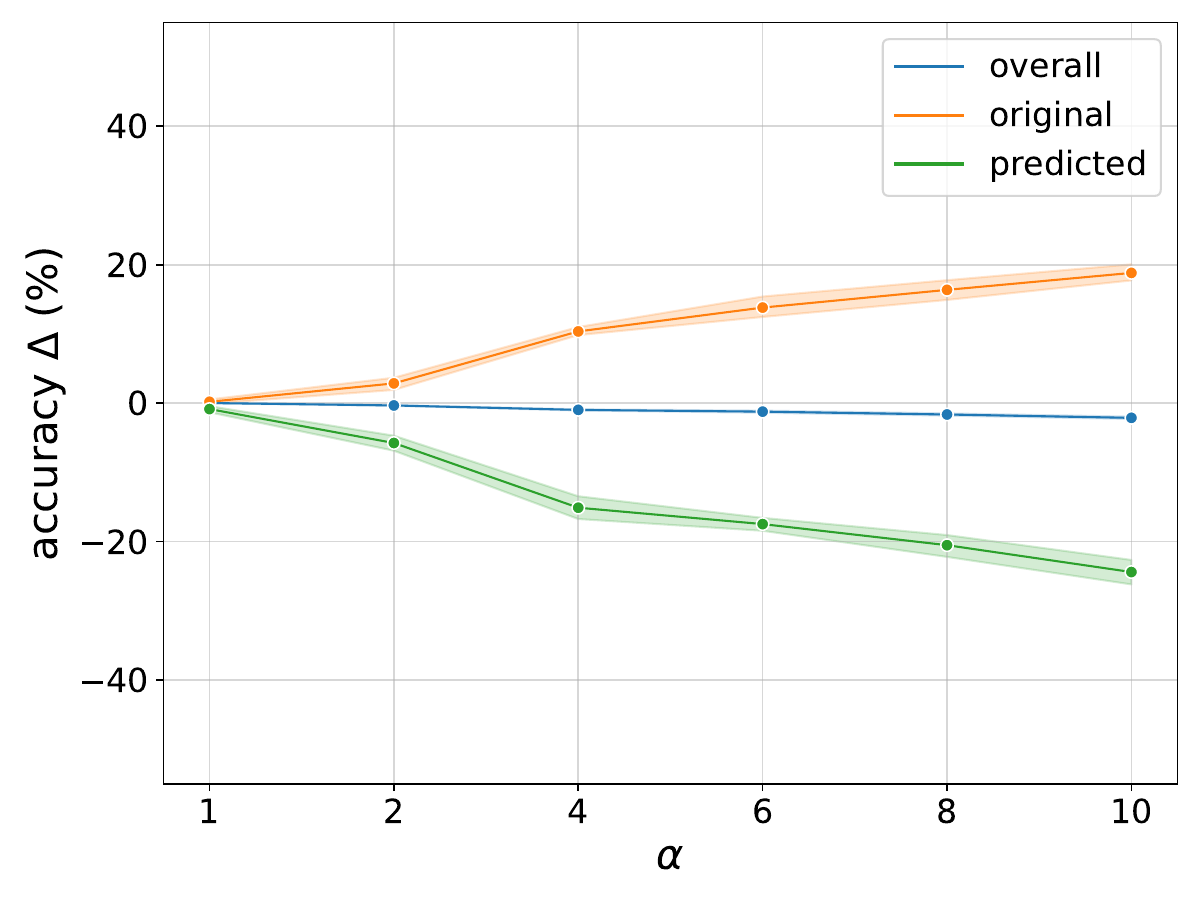}
        \caption{Fashion-MNIST (evaluation)}
        \label{fig:rq4_fm_eval}
    \end{subfigure}
    \vspace{0.5em}

    \begin{subfigure}{0.4\textwidth}
        \includegraphics[width=\textwidth, trim=0mm 5mm 0mm 0mm, clip]{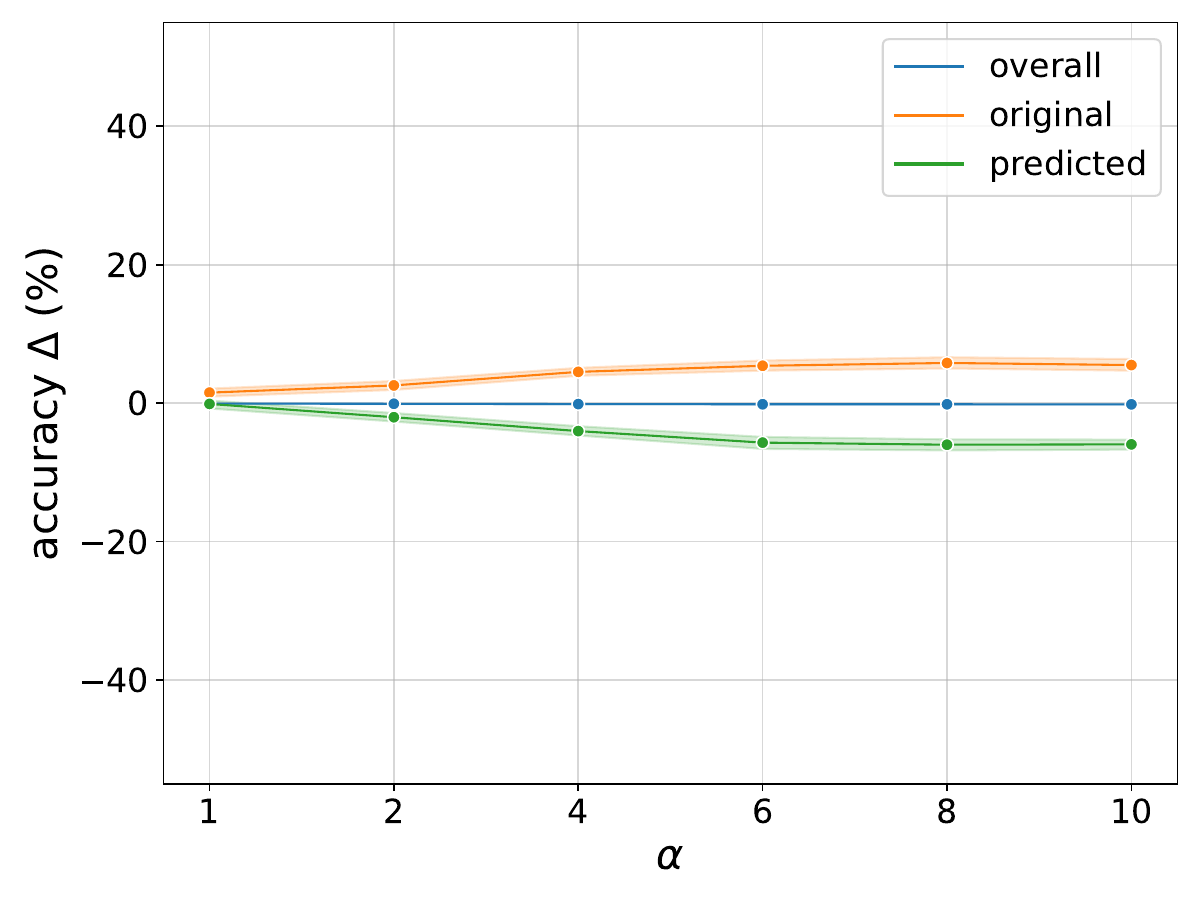}
        \caption{GTSRB (validation)}
        \label{fig:rq4_gtsrb_used}
    \end{subfigure}
    \begin{subfigure}{0.4\textwidth}
        \includegraphics[width=\textwidth, trim=0mm 5mm 0mm 0mm, clip]{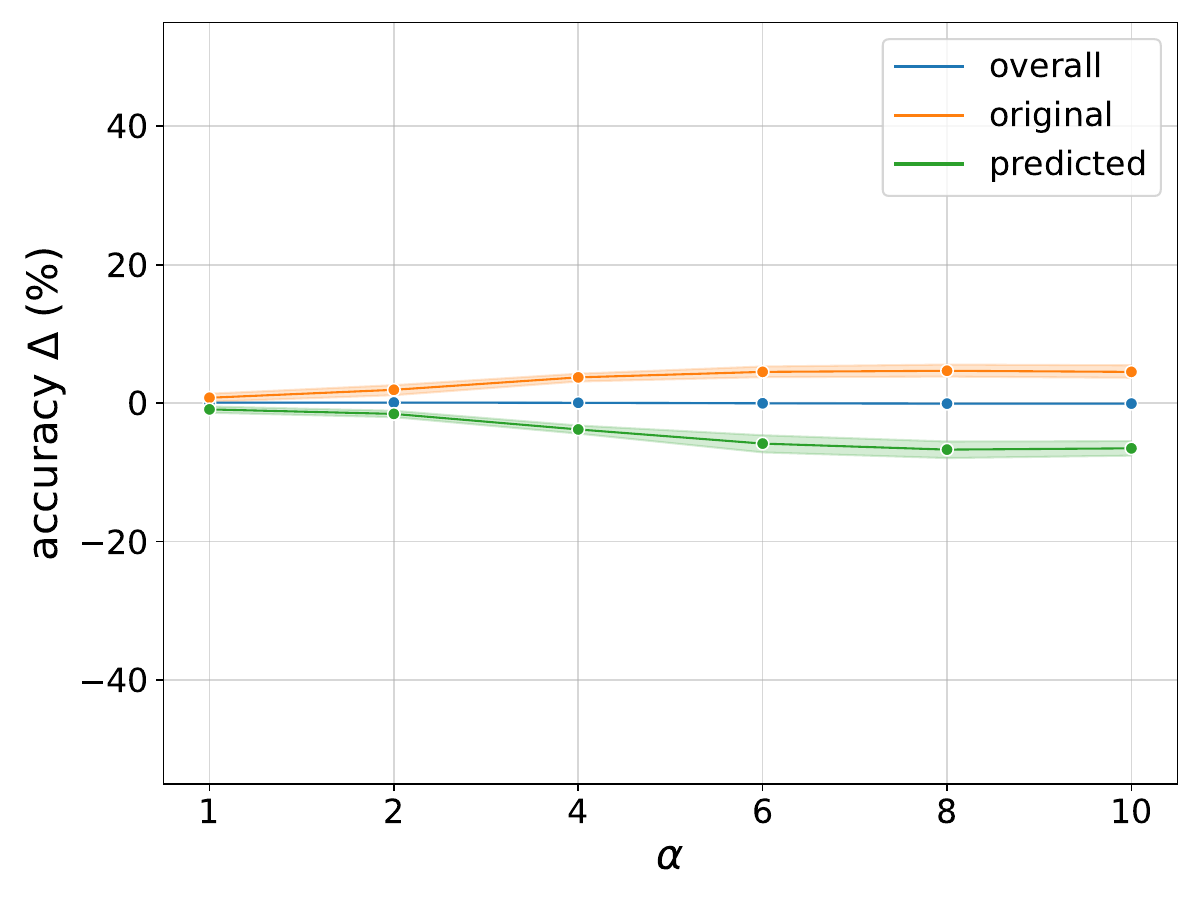}
        \caption{GTSRB (evaluation)}
        \label{fig:rq4_gtsbr_eval}
    \end{subfigure}

    \caption{The Impact of Using Different $\alpha$ on the most frequent types of misclassifications. The x-axis is $\alpha$ used in \name, and the y-axis ($\Delta$) is the corresponding degree of changes in accuracy.  
    The orange and green circles signify the results of the original class and the predicted class of the target misclassified inputs, respectively.
    The blue circles show how the overall accuracy is affected.
    }
    \label{fig:rq4}
\end{figure*}

\cref{fig:rq4} shows changing class accuracies and overall accuracy when the 
hyperparamter $\alpha$ is configured differently. The results reveal that 
increasing $\alpha$ does result in more aggressive repairs, even if it comes 
at the cost of breaking more existing behaviour. 
For instance, when $\alpha$ is set to 10 in CIFAR-10 (making the impact of patches most notable), \cref{fig:rq4} (b) shows that the patch increases the accuracy of the \corig class by 40\%p, at the cost of decreasing the accuracy of the \cpred class by 37\%p. 
\new{In contrast, when $\alpha$ is set to 1, \name neither fixes nor breaks as many inputs, %
repairing and breaking only 5.7\% and 3.8\% of the \corig and \cpred classes.
In Fashion MNIST, the overall trend is similar to that of CIFAR-10: with $\alpha=10$, and for the evaluation data (in \cref{fig:rq4} (d)), the accuracy of the \corig increases by 18.81\%p, whereas the accuracy of \cpred decreases by 24.4\%. When $\alpha=1$, the changes in both are within 1\%p. The same goes for the GTSRB, despite on a far smaller scale.}
In the end, although the degree of the impact is different for each model and dataset, the trend of varying $\alpha$ is to be consistent accross models (C10, FM, GTSRB) and datasets (validation and evaluation).\\

\noindent\fbox{\parbox{\textwidth}{
\textbf{Answer to RQ4:} We conclude that the hyperparameter $\alpha$ can 
effectively balance the behaviour of \name between aggressively pursuing 
repairs and conservatively retaining current behaviour.
}}

\subsection{RQ5: Comparison to the State-of-the-Art} %
\label{sub:rq5}

\begin{figure*}[ht!]
    \includegraphics[width=1.0\textwidth, trim=0mm 0mm 0mm 0mm, clip]{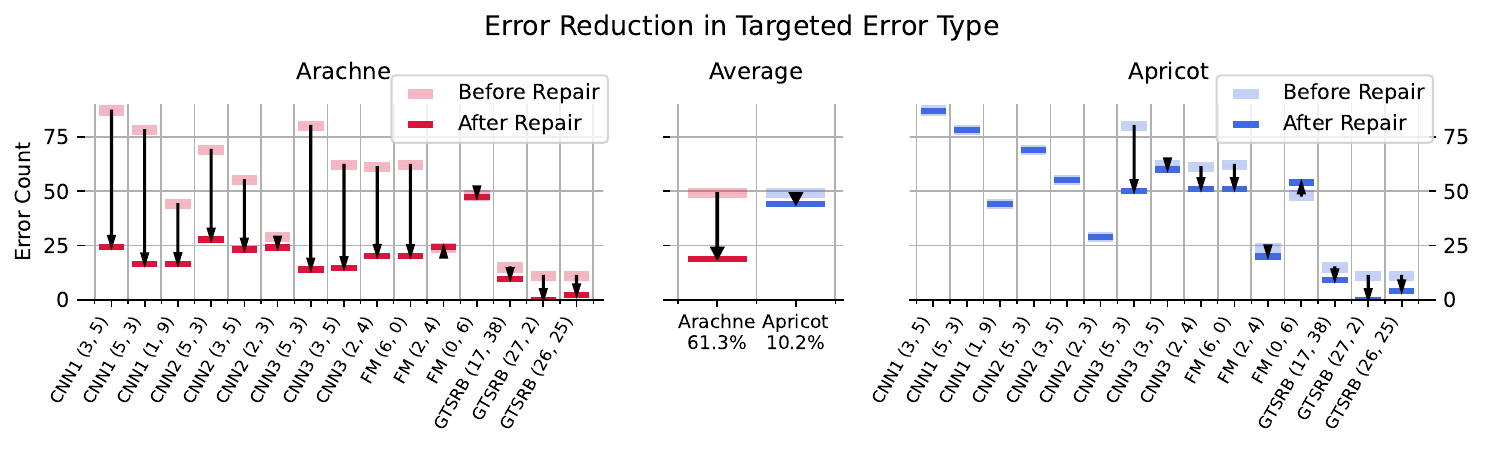}
    \caption{Number of wrong classifications of targeted type before and after repair, in \name and Apricot. The light bars represent the original errors of the targeted type, while the dark bars represent the average number of errors after the fix. These images were never seen by the neural network or the fixing algorithms. \minorrev{Retraining is omitted as it almost never changes performance.} Lower is better.
    \label{fig:rq5_sbs}}
\end{figure*}

\cref{fig:rq5_sbs} compares the number of reduced misclassifications per type before and after repairs performed by \name and Apricot. 
\name successfully reduces the error count of targeted misclassification types 
across all models: across all fifteen combinations of five image classifier models and three misclassification types, the number of misclassifications is reduced by $RR_{eval}=61.3$\%. 
\new{Apricot, on the other hand, is less successful than \name, repairing only $RR_{eval} = 10.2$\% of the targeted misclassifications on average.}

\new{Regarding the three image classifier models trained for CIFAR-10 (CNN1,2,3), \name successfully repairs the top three frequent misclassifications for all three models. In comparison, while Apricot does repair the misclassification for CNN3, the number of repaired errors is generally smaller than that of \name; for CNN1 and CNN2, Apricot does not report any change at all.}
We believe that the discrepancy in CNN1,2, and 3 between our results and those presented in Zhang and Chan~\cite{Zhang2019aa} is due to the different initial settings: we use DNNs trained until their performance on the test set plateau, while Zhang and Chan~\cite{Zhang2019aa} used DNNs trained until an arbitrary epoch threshold is met, making it possible that the DNNs in question were not fully trained.
\new{Meanwhile, in the FashionMNIST and GTSRB datasets that were not studied by Apricot, we observed that \name and Apricot perform similarly in generating targeted repairs in most cases except for the most frequent errors in FashionMNIST ($6 \rightarrow 0$), where \name outperformed Apricot; here, \name reduces the error count from 60 to near 20, whereas Apricot decreases it only to 50.}

\begin{table}[ht]
\centering
\caption{Average model accuracy of \name against the targeted version of Apricot and retraining (RT) for repairing frequent fault 1, 2, 3.
\label{tab:RQ5_acc}}
\scalebox{0.9}{
\begin{tabular}{ll|r|rrr|rrr|rrr}
\toprule
& & \textit{Initial} & \multicolumn{3}{c|}{\textit{Frequent Fault 1}} & \multicolumn{3}{c|}{\textit{Frequent Fault 2}} & \multicolumn{3}{c}{\textit{Frequent Fault 3}} \\ Model & & & \ns & Apri. & RT & \ns & Apri. & RT & \ns & Apri. & RT \\  \midrule

\multirow{2}{*}{\emph{CNN1}} & Train & \Numpfour{0.94274} & \Numpfour{0.914054} & \Numpfour{0.94274} & \minorrev{\Numpfour{0.923433}} & \Numpfour{0.906626} & \Numpfour{0.94274} & \minorrev{\Numpfour{0.923426}} & \Numpfour{0.9341853333333333} & \Numpfour{0.94274} & \minorrev{\Numpfour{0.92342}}\\
                            & Test  & \Numpfour{0.763} & \Numpfour{0.7519100000000001} & \Numpfour{0.763} & \minorrev{\Numpfour{0.763586}} & \Numpfour{0.7530800000000002} & \Numpfour{0.763} & \minorrev{\Numpfour{0.763586}} & \Numpfour{0.7622166666666669} & \Numpfour{0.763} & \minorrev{\Numpfour{0.763600}}\\

\midrule
\multirow{2}{*}{\emph{CNN2}} & Train & \Numpfour{1.0} & \Numpfour{0.9978293333333333} & \Numpfour{1.0} & \minorrev{\Numpfour{1.0}} &\Numpfour{0.9956026666666665} & \Numpfour{1.0} & \minorrev{\Numpfour{1.0}} & \Numpfour{0.9999533333333335} & \Numpfour{1.0} & \minorrev{\Numpfour{1.0}} \\
                            & Test  & \Numpfour{0.8336} & \Numpfour{0.82498} & \Numpfour{0.8336} & \minorrev{\Numpfour{0.83738}} & \Numpfour{0.822279999999999} & \Numpfour{0.8336} & \minorrev{\Numpfour{0.83742}} & \Numpfour{0.8299766666666667} & \Numpfour{0.8336} & \minorrev{\Numpfour{0.83741}} \\

\midrule
\multirow{2}{*}{\emph{CNN3}} & Train & \Numpfour{0.9956} & \Numpfour{0.9549366666666665} & \Numpfour{0.99962} & \minorrev{\Numpfour{0.995599}} & \Numpfour{0.9691853333333333} & \Numpfour{0.99966} & \minorrev{\Numpfour{0.985238}} & \Numpfour{0.9614246666666666} & \Numpfour{0.99948} & \minorrev{\Numpfour{0.974193}} \\
                            & Test  & \Numpfour{0.7638} & \Numpfour{0.743473333333333} & \Numpfour{0.7871} & \minorrev{\Numpfour{0.7638}} & \Numpfour{0.7524633333333334} & \Numpfour{0.778} & \minorrev{\Numpfour{0.756753}} & \Numpfour{0.73979} & \Numpfour{0.7759} & \minorrev{\Numpfour{0.7491066}}\\

\midrule
\multirow{2}{*}{\emph{FM}} & Train & \Numpfour{0.9948} & \Numpfour{0.9777605555555555} & \Numpfour{0.9997166666666667} & \minorrev{\Numpfour{0.9948}} & \Numpfour{0.9841344444444444} & \Numpfour{0.9998666666666667} & \minorrev{\Numpfour{0.9948}} & \Numpfour{0.99488} & \Numpfour{0.9999166666666667} & \minorrev{\Numpfour{0.9948}} \\
                            & Test  & \Numpfour{0.9267} & \Numpfour{0.91545} & \Numpfour{0.925} & \minorrev{\Numpfour{0.9267}} & \Numpfour{0.9212033333333333} & \Numpfour{0.9252} & \minorrev{\Numpfour{0.9267}} & \Numpfour{0.92694} & \Numpfour{0.9242} & \minorrev{\Numpfour{0.9267}}\\

\midrule
\multirow{2}{*}{\emph{GTSRB}} & Train & \Numpfour{1.0} & \Numpfour{0.9999379394186708} & \Numpfour{0.9999744956515085} & \minorrev{\Numpfour{1.0}} & \Numpfour{0.9998945820262354} & \Numpfour{0.9989288173633605} & \minorrev{\Numpfour{1.0}} & \Numpfour{0.999946440868168} & \Numpfour{0.9999744956515085} & \minorrev{\Numpfour{1.0}}\\
                              & Test  & \Numpfour{0.973396674584323} & \Numpfour{0.9730641330166272} & \Numpfour{0.9709962168978562} & \minorrev{\Numpfour{0.973396674584323}} & \Numpfour{0.975861704935339} & \Numpfour{0.967843631778058} & \minorrev{\Numpfour{0.973396674584323}} & \Numpfour{0.9732488783320137} & \Numpfour{0.967843631778058} & \minorrev{\Numpfour{0.973396674584323}}\\
\bottomrule
\end{tabular}}
\end{table}  

The high repair rate of \name may come at the cost of reduced overall 
accuracy, as \cref{tab:RQ5_acc} shows: when performing repair, \name may sacrifice overall accuracy to repair a specific type of misclassification. These results, combined with the higher repair rate, demonstrate that by construction \name is most useful as a targeted repair technique, and not as a retraining replacement for overall accuracy improvement. \minorrev{Meanwhile, we note that retraining barely changes performance; we believe this is due to the fact that while retraining uses gradient descent which can suffer when the optimization landscape is rough, \name uses the DE optimization technique which can overcome this issue.}

We further compare the time it takes to perform repair on these DNNs: 
\cref{tab:RQ5_time} shows the average repair time\footnote{In this experiment, we measure the entire running time of \name, from the localisation to the repair} for the Frequent Fault 1. 
\name does not require any additional training epochs. \new{On average, it terminates within ten minutes (i.e. two to seven mins.) for C10 and within three minutes for FM; For GTSRB, \name takes around average 23 minutes for the repair.}
In comparison, although we modified Apricot for early termination (reduced retraining epochs and early termination criterion: see Section~\ref{subsub:rq5}), %
\new{it takes nine (i.e., in GTSRB) to 86 times (i.e., in CNN3) longer than \name.} %
\new{These results further support the usefulness of Apricot in generating hot fixes that intend to quickly address errors of a certain type even at the cost of the others. %
}\\

\begin{table}[ht]
    \centering
    \caption{Average patch time of \name against Apricot for repairing Frequent Fault 1.
    \label{tab:RQ5_time}}
    \scalebox{1.0}{
    \begin{tabular}{l|r|r}
    \toprule
    model & \name & Apricot   \\  \midrule
    CNN 1 & 2m 6s & 1h 50m 12s  \\
    CNN 2 & 6m 8s & 2h 16m 58s  \\
    CNN 3 & 6m 41s & 9h 35m 52s  \\ 
    \new{F-MNIST} & \new{2m 19s} & \new{1h 36m 4s} \\
    \new{GTSRB} & \new{23m 19s} & \new{3h 29m 46s} \\
    \bottomrule
    \end{tabular}}
\end{table}

\noindent\fbox{\parbox{\textwidth}{
\textbf{Answer to RQ5:} we answer RQ5 by noting that \name outperforms the 
state-of-the-art DNN repair technique, Apricot, both in terms of \new{(targeted)} repair rate and execution time.
}}

\subsection{RQ6: Fairness Repair} %
\label{sub:rq6}

RQ6 concerns a case study of a realistic fairness repair for DNNs. The gender 
classification DNN model we train using the LFW benchmark~\cite{LFWTech,
afifi2019afif4} is biased due to the initial imbalance in the dataset: only 
22.41\% 
of the images in the LFW benchmark are those of female individuals.
As a result, the error rate of the female class among the test data is eight 
times higher than that of the male class, %
reaching \new{13.2\% against 2.7\%, respectively.}

\begin{figure}[ht!]
\centering
    \includegraphics[width=0.7\textwidth, trim=0mm 0mm 0mm 0mm, clip]{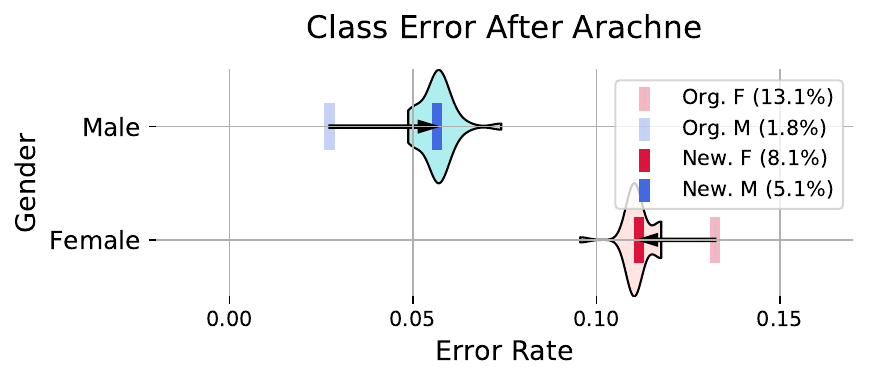}
    \caption{Class error before and after applying \name. Lower is better. The light contours represent the distribution of outcomes after 30 attempts to repair the neural network. The light bars indicate error by class before repair; the darker bars indicate average error after repair.\label{fig:rq6_genacc}}
\end{figure}

We apply \name on the classifier, collecting $I_{pos}$ and $I_{neg}$ from the holdout test set (i.e., validation set): $I_{pos}$ was all the images the classifier initially got correct, while $I_{neg}$ consisted of 18 misclassified \textit{female} images.
\cref{fig:rq6_genacc} shows how class accuracy changed when \name was applied.
After repair, the female error per class for the test data changed 
to 11.5\% and 5.6\% for the female and male classes, respectively.
The results suggest that we can use \name to address realistic fairness issues in DNN models, without resorting to data curation and retraining the model from scratch. We argue that issues such as model fairness can be an ideal application domain for \name, as quickly rebalancing the model may be necessary even if it degrades overall accuracy.

\noindent\fbox{\parbox{\textwidth}{
\textbf{Answer to RQ6:} \name can successfully repair DNNs with 
fairness issues, without additional data or retraining.}}

\subsection{RQ7: Model Generalisation} %
\begin{figure*}[t!]
    \centering
    \begin{subfigure}{0.415\textwidth}
        \includegraphics[width=\textwidth, trim=0mm 0mm 0mm 0mm, clip]{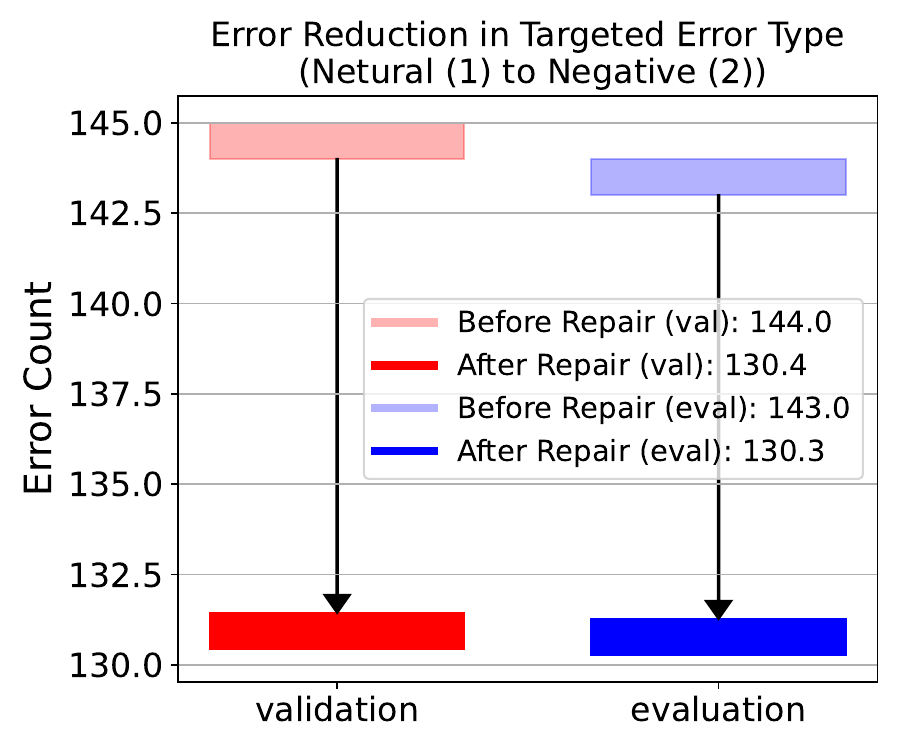} %
        \caption{
        \new{Error count reduction}
        \label{fig:rq7_change}}
    \end{subfigure} %
    \begin{subfigure}{0.5\textwidth}
        \includegraphics[width=\textwidth, trim=0mm 0mm 0mm 0mm, clip]{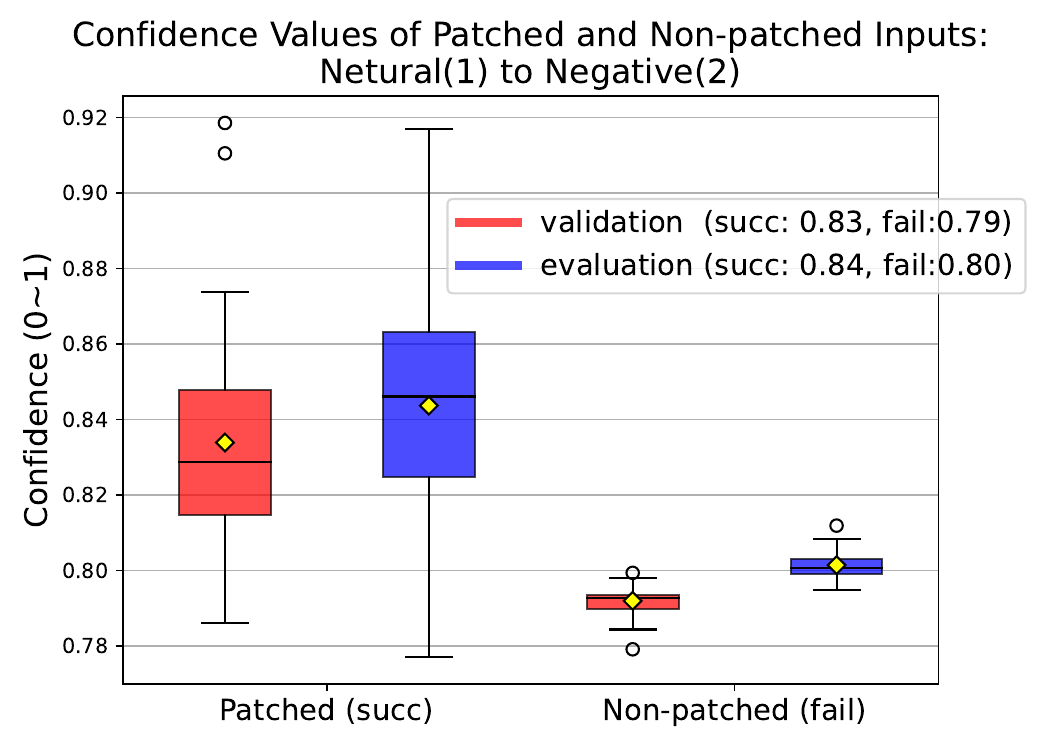}
        \caption{
        \new{Distribution of human evaluator confidence values}
        \label{fig:rq7_conf}}
    \end{subfigure}

    \caption{\new{Results from repairing the LSTM classifier for Twitter US aurline sentiment dataset. Blue and red denote the performance in the validation and evaluation set, respectively. The confidence values measure the confidence of human evaluator in their decision (positive/negative/neutral)}}
    \label{fig:rq7_combined}
\end{figure*}

\label{sec:rq7}
\new{
RQ7 considers whether \name can generate fixes for different data modalities 
and different DNN architectures than those considered in RQ1 through RQ6, 
namely image-classifying CNNs; thus as explained previously, we evaluate \name 
on a text dataset (i.e., the tweet dataset) and with an LSTM DNN. The results 
of this analysis are presented in \Cref{fig:rq7_change}, in which we target the 
most frequent misclassification type, i.e., the misclassification of negative 
tweets as neutral. We posit that this type of misclassification poses a higher 
risk than other misclassifications, as the model will potentially miss out on 
important customer feedback. 
As the figure shows, \name reduces the prevalence of this error type not only on the validation set (which is the basis for the adjusting of the DNN weights), but also on the evaluation set which was unseen to \name, suggesting the changes that \name makes indeed generalize to new inputs. To reemphasize, these results demonstrate that \name can operate over diverse data modalities and DNN architectures.

We further compare the confidence of labeler in the assigned label (included in 
the dataset) between inputs whose classifications are corrected by the patch generated by \name, and inputs whose classifications remain incorrect even after the patch. The results of this comparison are provided in \Cref{fig:rq7_conf}. We find that for the correctly patched inputs (i.e., inputs that were previously misclassified but now correctly classified), the confidence of human labelers is also high, whereas for the inputs \name fails to patch (i.e., inputs that continue to be misclassified despite the patch), the confidence of human labelers was low. 
These results suggest that \name repairs more worthwhile inputs, leaving only ambiguous cases misclassified. This provides further supporting evidence that 
\name can indeed generate worthwhile repairs for DNNs.
}
\\

\noindent\fbox{\parbox{\textwidth}{
\new{\textbf{Answer to RQ7:} \name can successfully repair completely different DNN architectures on different data modalities, suggesting the principles underpinning \name are general.}}}

\section{Threats to Validity}
\label{sec:threats}

Threats to the internal validity of our empirical evaluation include the 
stochasticity of \name, correctness of data collection, and the correctness of 
our implementation of Apricot. Since both \name and the training of DNN models are 
stochastic processes, randomness may affect the observed results. We repeat 
experiments whenever possible and report results from multiple types of 
misclassification to avoid this. 
We use widely studied frameworks for evolutionary computation and deep learning to build our models as well as reconstruct Apricot: DEAP~\cite{Fortin:2012aa} and PyTorch, respectively. We also make our implementation available for public scrutiny.

Threats to external validity concern any factors that may limit the 
generalisation of our claim. CIFAR-10 and Fashion-MNIST are widely studied 
benchmarks for image classification. \new{We also select GTSRB, the dataset that contains real-word traffic signs, to assess \name in a more practical case.} We augment these with the fairness repair study based on the LFW benchmark, with results that support the generalisability of our claims. 
\new{Lastly, we evaluate \name regarding the data modality and DNN architecture through the study other than the one on image classifier models, using a text dataset (i.e., the Twitter US airline sentiment dataset) and an LSTM DNN model.}
Additional studies can further strengthen the generalisability of our claims.

Threats to construct validity concern any potential misuse or 
misinterpretation of measured metrics. Our evaluation metrics are either 
standard classification metrics or simple count-based metrics 
with little room for misinterpretation.

\section{Related Work} %
\label{sec:related_work}

While there is a vast body of literature on techniques to improve the learning 
capability of DNNs~\cite{LeCun2015ef}, research on how to validate
the quality of DNN models is relatively young. Existing 
literature focus on ways to reveal misbehaviour by using the 
combination of input synthesis and metamorphic oracles~\cite{Pei2017qy,
Huang2017kx}. Several test adequacy criteria have also been introduced and 
studied~\cite{Pei2017qy,Tian2018aa, Kim2019aa,Ma2018aa}.

Debugging or patching a DNN model has so far been formulated in the context 
of (re)training, i.e., continuing to learn until the correct behaviour is 
trained. Ma et al. proposed MODE~\cite{Ma2018gf}, which uses Generative
Adversarial Networks (GANs) to synthesise additional inputs that focus on
the features relevant to the misbehaviour. These new inputs are used to
retrain the DNN model under repair. \name, on the other hand, formulates the
same problem as a direct Automated Program Repair (APR) for DNNs, and aims to 
solve it without generating or requiring additional data. 

Zhang and Chan proposed Apricot~\cite{Zhang2019aa}, which trains multiple 
``reduced Deep Learning Models'' (rDLMs) and incorporates them in an attempt 
to repair neural networks. Apricot trains multiple rDLMs, each of which have 
different classification results. By comparing the 
trained weights of rDLMs in these two different groups, Apricot readjusts the 
weights in the original DNN model towards correct rDLMs and applies additional training epochs.
Apricot is the closest to \name among the existing techniques, in that it does 
not require additional training data. 
However, unlike Apricot, \name does not rely on retraining at all.
Our results suggest that the additional 
training may be the primary source of its time cost.

\new{
Another approach towards retraining focuses on which additional input to use 
for the retraining. DeepFault~\cite{Eniser2019qr}, for example, identifies 
\emph{suspicious neurons} using a spectrum based approach, and perturbs 
existing input images in the direction that increases the activation value of 
the suspicious neurons. As a result, DeepFault can exploit the suspicious 
neurons (neurons that are correlated with DNN misbehaviour) to synthesize 
adversarial examples, which can subsequently be used for retraining. However, 
due to the way DeepFault synthesises new inputs based on domain constraints, 
its application is currently limited to Convolutional Neural Networks, whereas 
\name can be applied to any DNN architecture.}

\name is heavily inspired by a class of APR techniques called Generate and 
Validate (G\&V)~\cite{LeClair2012,Goues:2012zr,Yuan2018zl,Wen2018dk}. Concepts
such as the use of positive and negative input sets, the use of fault 
localisation, and the use of metaheuristic search as the main driver of the 
repair are all inherited from existing G\&V techniques. \new{Regarding fault 
localisation, DeepFault~\cite{Eniser2019qr} adopts the Spectrum Based Fault 
Localisation (SBFL) framework that has been successfully applied to coverage 
and test result information from traditional programs~\cite{Wong:2016aa}. 
\minorrev{However, unlike \name that localises neural weights, DeepFault localises neurons. 
This coarser granularity of DeepFault requires an additional process to 
identify true-faulty neural weights among those connected to localised neurons 
for it to be used in the patch generation phase of \name. While treating all 
connected neural weights of faulty neurons as places to fix can be another option, 
this may lead to search space explosion.}
Further, DeepFault requires the user to set an arbitrary threshold for 
neural activation value to fit DNN activations into the program spectrum 
framework. \name does not require such hyperparameters, as its localisation 
depends on both gradient loss and forward impact of each neuron weight.} 

Regarding patching, like other G\&V 
techniques, \name first \emph{generates} a candidate patch and \emph{validates}
the patch by applying it and subsequently executing the patched model against 
the set of positive and negative inputs. However, due to the nature of DNNs,
some fundamental differences exist. Both the program and the patch 
representation are numerical and continuous for \name. Unlike the fitness 
function of GenProg~\cite{LeClair2012,Goues:2012zr} that only counts discrete 
step changes in the number of test cases that fail, \name can use the model 
loss as a continuous guidance towards repair. \new{While this paper presents 
separate studies for accuracy and fairness repair, it is possible to repair a 
given DNN with respect to multiple objectives, for example by following the multi-objective formulation of APR introduced by ARJA~\cite{Yuan2018zl}. This is possible because \name does not require any repair-specific modifications to the DNN.}

\section{Conclusion}
\label{sec:conclusion}

We present \name, a search-based repair technique for DNN models. \name uses 
Differential Evolution to directly manipulate neural weight values, which are 
chosen by a novel localisation method. An empirical evaluation 
using widely studied \new{four image benchmark datasets and one text dataset},  suggests that \name can 
repair misbehaviour of DNN models while minimally disrupting existing 
correct behaviour. Furthermore, in targeted repair, \name outperforms a 
state-of-the-art method and tens of times faster. %
We also conduct a case study of realistic fairness repair: \name can repair a biased gender classification model by rebalancing the neural weights without 
requiring any additional data. \new{The additional study with a text dataset and an LSTM-based model further supports the applicability of \name in different domains of datasets and DNN architectures.}
We believe that \name will facilitate the deployment of DNNs, acting as a fast 
adjustment technique capable of fixing both fairness and misclassification 
issues.

\bibliographystyle{IEEEtran}
\bibliography{newref}

\end{document}